\pdfoutput=1

\documentclass[a4paper,12pt]{article}

\newif\ifdraft
\draftfalse

\newif\ifword
\wordfalse

    \usepackage{abstract} %

    \usepackage{titlesec} %

    \usepackage{authblk} %

	\usepackage{datetime} %
    \newdateformat{usvardate}{
  	\monthname[\THEMONTH] \ordinal{DAY}, \THEYEAR}

  	\usepackage[bottom]{footmisc} %

 \ifword
 \else
    \usepackage{fancyhdr}
    \fancyheadoffset{0cm}
 \fi

\usepackage[section]{placeins}

    \usepackage[USenglish]{babel}
\newcommand\wordcount{%
  \immediate\write18{texcount -utf8 -merge -sum -incbib -dir -sub=none -brief \jobname.tex | cut -d : -f 1 > 'count.txt'}%
  \input{count.txt}\ignorespaces words%
}

    \usepackage{amssymb}
    \usepackage{amsthm}
    \usepackage{amsmath,amsfonts}

    \usepackage{siunitx} %

    \usepackage[super]{nth}
    
    \usepackage{gensymb}

    \usepackage[utf8]{inputenc}

	\ifword
        
		\usepackage{fontspec}

		\usepackage[cmintegrals,cmbraces,vvarbb]{newtxmath}
		\usepackage[none]{hyphenat}
		\defaultfontfeatures{Ligatures=NoCommon}
	\else
		\usepackage{libertine} %
		\usepackage[libertine,cmintegrals,cmbraces,vvarbb]{newtxmath} %
        
        \usepackage[activate={true,nocompatibility},final,tracking=true,kerning=true,spacing=true,factor=1100,stretch=10,shrink=10,babel]{microtype}
	\fi

	\usepackage{lettrine}
	\usepackage{setspace} %
    \usepackage{lineno,xcolor} 	%

    \usepackage{multicol} %

    \usepackage{enumitem} %

	\usepackage[top=3cm, bottom=3cm, left=2.5cm, right=3cm]{geometry}
	\ifdraft
	  \usepackage[colorinlistoftodos,shadow,loadshadowlibrary]{todonotes} %
	\else
	  \usepackage[disable,colorinlistoftodos,shadow,loadshadowlibrary]{todonotes} %
	\fi
	\usepackage{soul} %

    \usepackage{graphicx} %
    \usepackage{float} %

    \usepackage{printlen}

	\usepackage{booktabs} %
    \usepackage{longtable} %
    \usepackage{tabularx} %
    \usepackage{multirow} %
    \usepackage{rotating} %

    \newcolumntype{L}[1]{>{\raggedright\arraybackslash}p{#1}}
    \newcolumntype{C}[1]{>{\centering\arraybackslash}p{#1}}
    \newcolumntype{R}[1]{>{\raggedleft\arraybackslash}p{#1}}
    \usepackage{relsize} %
	\usepackage{lscape} %
	\usepackage{caption}
	\usepackage{acro} %

	\newlist{acronyms}{description}{1}
	\newcommand *\addcolon[1]{#1}
	\setlist[acronyms]{
		labelwidth = 3em,
		leftmargin = 1cm,
		rightmargin= .7cm,
		noitemsep,
		itemindent = 0pt,
		font=\addcolon\small}

	\definecolor{MUW Dunkelblau}{RGB}{17,29,78}
	\definecolor{MUW Hellblau 1}{RGB}{95,180,229}
	\definecolor{MUW Hellblau 2}{RGB}{151,207,236}
	\definecolor{MUW Hellblau 3}{RGB}{181,220,241}
	\definecolor{MUW Hellblau 4}{RGB}{207,232,246}
	\definecolor{MUW Hellblau 5}{RGB}{231,243,251}
	\definecolor{MUW Gruen 1}{RGB}{47,142,145}
	\definecolor{MUW Gruen 2}{RGB}{132,201,188}
	\definecolor{MUW Gruen 3}{RGB}{168,215,205}
	\definecolor{MUW Gruen 4}{RGB}{200,229,223}
	\definecolor{MUW Gruen 5}{RGB}{228,242,239}
	\definecolor{MUW Skin 1}{RGB}{240,167,148}
	\definecolor{MUW Skin 2}{RGB}{248,192,176}
	\definecolor{MUW Skin 3}{RGB}{250,209,196}
	\definecolor{MUW Skin 4}{RGB}{252,225,216}
	\definecolor{MUW Skin 5}{RGB}{253,240,235}
	\definecolor{MUW Skin 6}{RGB}{254,247,245}
	\definecolor{RET Green 1}{RGB}{175,202,11}
    \definecolor{RET Green 2}{RGB}{134,188,37}
    \definecolor{RET Green 3}{RGB}{58,170,53}
    \definecolor{RET Blue 1}{RGB}{0,142,207}
    \definecolor{RET Blue 2}{RGB}{73,100,140}

	\usepackage[breaklinks=true]{hyperref} %
	\hypersetup{
		colorlinks,
		citecolor=RET Blue 1,
		linkcolor=RET Green 3,
		urlcolor=RET Blue 2}

    \usepackage[super]{natbib}

    \makeatletter
    \renewcommand\@biblabel[1]{#1\quad}
    \makeatother

\ifdraft
    \usepackage[draft,truncate=hyphenate,commentmarkup=uwave]{changes} %
\else
    \usepackage[final,truncate=hyphenate,commentmarkup=uwave]{changes} %
\fi

\definechangesauthor[name={Wolf-Dieter}, color=RET Green 1]{W}
\definechangesauthor[name={Ariadne}, color=RET Blue 1]{A}
\definechangesauthor[name={Oliver}, color=RET Blue 2]{O}

\ifword

\else
\fi

\DeclareAcronym{GA}{short = GA , long = geographic atrophy}
\DeclareAcronym{LPR}{short = LPR , long = local progression rate}
\DeclareAcronym{FAF}{short = FAF , long = fundus autofluorescence}
\DeclareAcronym{EOM}{short = EOM , long = every other month}
\DeclareAcronym{SM}{short = SM , long = Sham}
\DeclareAcronym{PR}{short = PR , long = photoreceptor}
\DeclareAcronym{HRF}{short = HRF, long = hyperreflective foci}
\DeclareAcronym{pp}{short = pp, long = percentage points}
\DeclareAcronym{DL}{short = DL, long = deep learning}
\DeclareAcronym{CC}{short = CC, long = cross correlation}
\DeclareAcronym{PDE}{short = PDE, long = partial differential equation}
\DeclareAcronym{GLMM}{short = GLMM, long = generalized linear mixed model}
\DeclareAcronym{GAMM}{short = GAMM, long = generalized additive mixed model}
\DeclareAcronym{GAM}{short = GAM, long = generalized additive model}
\DeclareAcronym{iid}{short = i.i.d, long = independent and identically distributed}
\DeclareAcronym{PQL}{short = PQL, long = penalized quasi-likelihood}
\DeclareAcronym{CNN}{short = CNN, long = convolutional neural network}
\DeclareAcronym{SD-OCT}{short = SD-OCT , long = spectral domain optical coherence tomography}
\DeclareAcronym{EZ}{short = EZ, long = ellipsoid zone}
\DeclareAcronym{IB-EZ}{short = IB-EZ, long = inner boundary of the ellipsoid zone}
\DeclareAcronym{OB-OPR}{short = OB-OPR, long = outer boundary of outer photoreceptors}
\DeclareAcronym{CE}{short = CE, long = cross entropy}
\DeclareAcronym{SGD}{short = SGD, long = stochastic gradient descent}
\DeclareAcronym{FPN}{short = FPN, long = Feature Projection Network}
\DeclareAcronym{ASSD}{short = ASSD, long = average symmetric surface distance}
\DeclareAcronym{NPV}{short = NPV, long = negative predictive value}
\DeclareAcronym{HD}{short = HD, long = Hausdorff distance}
\DeclareAcronym{HD95}{short = HD95, long = Hausdorff distance \nth{95} percentile}
\DeclareAcronym{IQR}{short = IQR, long = interquartile range}
\DeclareAcronym{TIMM}{short = TIMM, long = PyTorch Image Models}
\DeclareAcronym{ICC}{short = ICC, long = intraclass correlation coefficients}
\DeclareAcronym{SQRT}{short = SQRT, long = square-root transformed}
\DeclareAcronym{AD}{short = AD, long = absolute deviation}
\DeclareAcronym{PD}{short = PD, long = absolute percentage deviation}
\DeclareAcronym{MAPE}{short = MAPE, long= mean absolute percentage error}
\DeclareAcronym{DR}{short = DR, long= Deming regression}
\DeclareAcronym{SDD}{short = SDD, long= subretinal drusenoid deposits}
\DeclareAcronym{GT}{short = GT, long=ground-truth}
\DeclareAcronym{LoA}{short = LoA, long=limit of agreement, long-plural=limits of agreement}
\DeclareAcronym{iRORA}{short = iRORA, long=incomplete RPE and outer retinal atrophy}
\DeclareAcronym{MA}{short = MA , long = macular atrophy}
\DeclareAcronym{MNV}{short = MNV , long = macular neovascularization}
\DeclareAcronym{PED}{short = PED , long = pigment epithelial detachment}
\DeclareAcronym{CLAHE}{short = CLAHE, long = contrast limited adaptive histogram equalization}
\DeclareAcronym{RNFL}{short = RNFL, long = retinal nerve fibre layer}
\DeclareAcronym{GCL}{short = GCL, long = ganglion cell layer}
\DeclareAcronym{IPL}{short = IPL, long = inner plexiform layer}
\DeclareAcronym{INL}{short = INL, long = inner nuclear layer}
\DeclareAcronym{OPL}{short = OPL, long = outer plexiform layer}
\DeclareAcronym{ONL}{short = ONL, long = outer nuclear layer}
\DeclareAcronym{ORB}{short = ORB, long = outer retinal hyperreflective bands}
\DeclareAcronym{FDR}{short = FDR, long = false discovery rate}
\DeclareAcronym{std}{short = std, long = standard deviation}
\DeclareAcronym{LOESS}{short = LOESS, long = locally estimated scatterplot smoothing}
\DeclareAcronym{DOM}{short = DOM, long = difference of means}
\DeclareAcronym{SNR}{short = SNR, long = signal to noise ratio}
\DeclareAcronym{PR-ISOS}{short = PR-IS/OS, long = photoreceptor-inner segment/outer segment}
\DeclareAcronym{BM}{short = BM, long = Bruch's membrane}
\DeclareAcronym{HFL}{short = HFL, long = Henle fiber layer}
\DeclareAcronym{SS-OCT}{short = SS-OCT, long = swept-source OCT}
\DeclareAcronym{OCT-A}{short = OCT-A, long = OCT angiography}
\DeclareAcronym{AO-OCT}{short = AO-OCT, long = adaptive optics OCT}
\DeclareAcronym{D-OCT}{short = D-OCT, long = directional OCT}
\DeclareAcronym{SHRM}{short = SHRM, long = subretinal hyperreflective material}

\DeclareAcronym{ONH}{short = ONH , long = optic nerve head}
\DeclareAcronym{OCT}{short = OCT , long = optical coherence tomography}
\DeclareAcronym{ILM}{short = ILM , long = inner limiting membrane}
\DeclareAcronym{RPE}{short = RPE , long = retinal pigment epithelium}
\DeclareAcronym{GMM}{short = GMM , long = Gaussian mixture model}
\DeclareAcronym{EM}{short = EM , long = expectation maximization}
\DeclareAcronym{SLO}{short = SLO , long = scanning laser ophthalmoscope}
\DeclareAcronym{GLM}{short = GLM , long = generalized linear model}
\DeclareAcronym{MAP}{short = MAP , long = maximum a posteriori probability}
\DeclareAcronym{CMA-ES}{short = CMA-ES , long = covariance matrix adaptation evolutionary strategy}
\DeclareAcronym{RF}{short = RF , long = random forests}
\DeclareAcronym{ET}{short = ET , long = extra trees}
\DeclareAcronym{CRVO}{short = CRVO , long = central retinal vein occlusion}
\DeclareAcronym{BRVO}{short = BRVO , long = branch retinal vein occlusion}
\DeclareAcronym{AuC}{short = AuC , long = area under ROC curve}
\DeclareAcronym{OOB}{short = OOB , long = out-of-bag}
\DeclareAcronym{CI}{short = CI , long = confidence interval}
\DeclareAcronym{TRT}{short = TRT , long = total retinal thickness}
\DeclareAcronym{CV}{short = CV , long = cross-validation}
\DeclareAcronym{RVO}{short = RVO , long = retinal vein occlusion}

\DeclareAcronym{AMD}{short = AMD , long = age-related macular degeneration}
\DeclareAcronym{iAMD}{short = iAMD , long = intermediate AMD}
\DeclareAcronym{nAMD}{short = nAMD , long = neovascular AMD}
\DeclareAcronym{PCA}{short = PCA , long = principal component analysis}
\DeclareAcronym{ICA}{short = ICA , long = independent component analysis}
\DeclareAcronym{MDI}{short = MDI , long = mean decrease in impurity}
\DeclareAcronym{LASSO}{short = LASSO , long = least absolute shrinkage and selection operator}
\DeclareAcronym{OLS}{short = OLS , long = ordinary least squares}
\DeclareAcronym{GLS}{short = GLS , long = generalized least squares}
\DeclareAcronym{RLS}{short = RLS , long = regularized least squares}
\DeclareAcronym{BLUE}{short = BLUE , long = best linear unbiased estimator}
\DeclareAcronym{EBLUE}{short = EBLUE , long = empirical best linear unbiased estimator}
\DeclareAcronym{MLE}{short = MLE , long = maximum likelihood estimation}
\DeclareAcronym{ML}{short = ML , long = maximum likelihood}
\DeclareAcronym{REML}{short = REML , long = restricted maximum likelihood}
\DeclareAcronym{MRM}{short = MRM , long = mixed effects regression model}
\DeclareAcronym{MCMC}{short = MCMC , long = Markov chain Monte Carlo}
\DeclareAcronym{EVD}{short = EVD , long = eigenvalue decomposition}
\DeclareAcronym{SVD}{short = SVD , long = singular value decomposition}
\DeclareAcronym{LRT}{short = LRT , long = likelihood ratio test}
\DeclareAcronym{AIC}{short = AIC , long = Akaike information criterion}
\DeclareAcronym{BIC}{short = BIC , long = Bayesian information criterion}
\DeclareAcronym{BLUP}{short = BLUP , long = best linear unbiased predictor}
\DeclareAcronym{EBLUP}{short = EBLUP , long = empirical BLUP}
\DeclareAcronym{EB}{short = EB , long = empirical Bayes}
\DeclareAcronym{SyGN}{short = SyGN , long = symmetric group-wise normalization}
\DeclareAcronym{CT}{short = CT , long = computed tomography}
\DeclareAcronym{LDDMM}{short = LDDMM , long = large deformation diffeomorphic metric mapping}
\DeclareAcronym{CPD}{short = CPD , long = coherent point drift}
\DeclareAcronym{ANTs}{short = ANTs , long = advanced normalization tools}
\DeclareAcronym{ode}{short = o.d.e. , long = ordinary differential equation ,short-plural='s}
\DeclareAcronym{SyN}{short = SyN , long = symmetric normalization}
\DeclareAcronym{MSQ}{short = MSQ , long = mean squared difference}
\DeclareAcronym{NCC}{short = NCC , long = normalized cross-correlation}
\DeclareAcronym{MI}{short = MI , long = mutual information}
\DeclareAcronym{SSD}{short = SSD , long = sum of squared differences}
\DeclareAcronym{KL}{short = KL , long = Kullback-Leibler}
\DeclareAcronym{FILD}{short = FILD , long = fibrosing interstitial lung disease}
\DeclareAcronym{GLCM}{short = GLCM , long = gray-level co-occurence matrix}
\DeclareAcronym{BCD}{short = BCD , long = Bray-Curtis dissimilarity}
\DeclareAcronym{DSC}{short = DSC , long = Dice similarity coefficient}
\DeclareAcronym{SLSQP}{short = SLSQP , long = sequential least squares programming}
\DeclareAcronym{SQP}{short = SQP , long = sequential quadratic programming}
\DeclareAcronym{HRCT}{short = HRCT , long = high resolution computed tomography}
\DeclareAcronym{UIP}{short = UIP , long = usual interstitial pneumonia}
\DeclareAcronym{NSIP}{short = NSIP , long = nonspecific interstitial pneumonia}
\DeclareAcronym{EAA}{short = EAA , long = extrinsic allergic alveolitis}
\DeclareAcronym{LOOCV}{short = LOOCV , long = leave-one-out cross-validation}
\DeclareAcronym{ROC}{short = ROC , long = receiver-operating-characteristic}
\DeclareAcronym{BSL}{short = BSL , long = baseline}
\DeclareAcronym{ETDRS}{short = ETDRS , long = early treatment diabetic retinopathy study}
\DeclareAcronym{CRT}{short = CRT , long = central retinal thickness}
\DeclareAcronym{BCVA}{short = BCVA , long = best corrected visual acuity}
\DeclareAcronym{VA}{short = VA , long = visual acuity}
\DeclareAcronym{IRF}{short = IRF , long = intraretinal fluid}
\DeclareAcronym{SRF}{short = SRF , long = subretinal fluid}
\DeclareAcronym{MAE}{short = MAE , long = mean absolute error}
\DeclareAcronym{SD}{short = SD , long = standard deviation}
\DeclareAcronym{SE}{short = SE , long = standard error}
\DeclareAcronym{DM}{short = DM , long = Diebold-Mariano}
\DeclareAcronym{IIP}{short = IIP, long = idiopathic interstitial pneumonia}
\DeclareAcronym{IPF}{short = IPF, long = idiopathic pulmonary fibrois}
\DeclareAcronym{VEGF}{short = VEGF, long = vascular endothelial growth factor}
\DeclareAcronym{prn}{short = PRN, long = pro re nata}
\DeclareAcronym{GEE}{short = GEE, long = generalized estimating equation}
\DeclareAcronym{fmri}{short = fMRI, long = functional magnetic resonance imaging}
\DeclareAcronym{ANOVA}{short = ANOVA, long = analysis of variance}

\DeclareMathAlphabet{\mathbfsf}{\encodingdefault}{\sfdefault}{bx}{n}

\begin{document}

  \title{Fully Automated Deep Learning Segmentation of RPE Loss and Ellipsoid Zone Integrity in Geographic Atrophy on Real-World OCT}

  \author[1,*]{Wolf-Dieter Vogl}
  \author[1]{Hlynur Skulason}
  \author[1]{Oliver Leingang}
  \author[2]{Ursula Schmidt-Erfurth}
  \author[1]{Amir Sadeghipour}
  \author[1]{Ariadne Whitby}

  \affil[1]{RetInSight GmbH, Vienna, Austria}
  \affil[2]{Laboratory of Ophthalmic Image Analysis (OPTIMA), Medical University of Vienna, Austria}
  \affil[*]{wolf-dieter.vogl@retinsight.com}

  \renewcommand\Authands{ and }

	\date{}

\maketitle

\clearpage

\ifword

\else
  \renewcommand\linenumberfont{\normalfont\footnotesize\sffamily\color{gray}}
\fi

\begin{abstract}
\textbf{Purpose:} To develop and validate a fully automated deep learning method for segmenting retinal pigment epithelium (RPE) loss, ellipsoid zone (EZ) loss, and EZ thickness in geographic atrophy (GA) on optical coherence tomography (OCT).

\textbf{Methods:} Three convolutional neural networks segment Bruch's membrane, the EZ boundaries, and RPE loss in Spectralis OCT volumes; EZ loss and thickness are derived from the EZ boundaries. The networks were developed on 298 volumes (GA, intermediate and neovascular age-related macular degeneration, healthy) and evaluated against double-graded expert annotations in an independent real-world set of 43 eyes with GA. Further analyses assessed inter-reader reliability (25 eyes), test--retest repeatability between screening and baseline visits of two phase 3 trials (68 eyes), and the effect of B-scan density (61 eyes).

\textbf{Results:} Mean Dice similarity coefficients were 0.88 (95\% confidence interval [CI], 0.84--0.90) for RPE loss and 0.87 (95\% CI, 0.83--0.89) for EZ loss; the mean absolute EZ thickness error was 2.08~\si{\micro\metre}. Agreement of the algorithm with an independent reader group was similar to inter-reader agreement. Test--retest intraclass correlation coefficients were 0.995 (RPE loss) and 0.988 (EZ loss). Lower B-scan densities had little effect on lesion area but markedly increased the error in measuring 3-month change.

\textbf{Conclusions:} On real-world OCT, the fully automated method was accurate, with agreement comparable to human readers, and reliable, with high test--retest repeatability.

\textbf{Translational Relevance:} Fully automated, reproducible quantification of RPE and photoreceptor loss on routine OCT could support standardized structural endpoints in GA clinical trials and objective assessment of GA in routine ophthalmic care.
\end{abstract}

\section{Introduction}
\lettrine[lines=2]{A}{ge-related macular degeneration} (\acs{AMD}) remains a leading cause of irreversible visual impairment worldwide, particularly within aging demographics~\cite{chakravarthyCharacterizingDiseaseBurden2018}. \Ac{GA} represents the advanced "dry" stage of AMD, characterized by progressive loss of \ac{RPE}, overlying \acp{PR}, and the underlying choriocapillaris~\cite{holzGeographicAtrophyClinical2014}. The expanding lesions compromise the macula’s functional integrity, leading to scotoma and irreversible practical blindness. In clinical practice, \ac{OCT} has meanwhile superseded fundus autofluorescence as the gold standard for diagnosing and monitoring GA~\cite{reiterAIClinicalManagement2024}. By providing high-resolution, cross-sectional visualizations of retinal microlayers, \ac{OCT} enables a precise demarcation of atrophic boundaries. In \ac{OCT} imaging, \ac{GA} is primarily identified by the attenuation of \ac{PR}-related structures, \ac{RPE} degeneration, and a subsequent increase in choroidal hypertransmission~\cite{saddaConsensusDefinitionAtrophy2018}. 

Functional decline in \ac{GA} is primarily driven by \ac{PR} atrophy. While \ac{RPE} loss is often the most prominent clinical marker, \ac{PR} survival depends strongly on the metabolic support provided by the underlying RPE. Recent longitudinal studies have highlighted that photoreceptor thinning and disruption consistently precede the loss of the RPE layer~\cite{zekavatPhotoreceptorLayerThinning2022}.
In morphological imaging, the \ac{EZ} is the primary OCT biomarker for monitoring \ac{PR} integrity. The \ac{EZ} appears as a hyperreflective layer, located immediately above the RPE, representing the mitochondria-rich ellipsoid of the photoreceptor inner segments~\cite{staurenghiProposedLexiconAnatomic2014}. \ac{EZ} thickness and integrity has been shown to be a highly relevant biomarker correlating with retinal function, disease progression rate, and treatment response~\cite{reiterSubretinalDrusenoidDeposits2020, pfauProgressionPhotoreceptorDegeneration2020, riedlEffectPegcetacoplanTreatment2022, voglPredictingTopographicDisease2023, maresCorrelationRetinalFluid2025, birnerStructureFunctionCorrelationDeepLearning2025, birnerNormativeProspectiveData2025, maiDynamicsEZRPE2025}. EZ-based measures are now also used as primary endpoints in phase 3 trials, for example the rate of change in the macular area of photoreceptor loss in the ReNEW trial of elamipretide in dry AMD~\cite{stealthbiotherapeuticsinc.ReNEWPhase32025}.

Manual segmentation of \ac{GA} lesions and the quantification of \ac{EZ} integrity are labor-intensive tasks, rendering the process impractical for routine clinical workflows. Manual segmentation of a pathological and low-reflectance layer is also affected by subjective variability. Recent advancements in automated \ac{DL} algorithms - specifically \acp{CNN} - have enabled high-throughput, objective segmentation of the biomarkers \ac{RPE} loss~\cite{ruiz-morenoAutomaticQuantificationSoftware2020, lachinovProjectiveSkipConnectionsSegmentation2021,  pramilDeepLearningModel2023, moranoDeepMultimodalFusion2024, spaideEstimatingUncertaintyGeographic2025, al-khersanDeepLearningBasedSegmentation2025} and \ac{EZ} degradation~\cite{orlandoAutomatedQuantificationPhotoreceptor2020, pfauProgressionPhotoreceptorDegeneration2020, kalraAutomatedIdentificationSegmentation2023, schmidt-erfurthDiseaseActivityTherapeutic2025, birnerExploringTrialEndpoints2026}.
However, a significant limitation is that many of these models are trained on curated clinical trial data with strict inclusion and exclusion criteria, such as specific lesion size ranging. This reliance on "clean" data may compromise the generalizability and performance of these models when applied to diverse, real-world clinical populations~\cite{suNavigatingDistributionShifts2025}. 

In this work, we present a fully automated \ac{DL} framework for the pixel-wise segmentation of \ac{RPE} loss, \ac{EZ} layer loss, and \ac{EZ} thinning using an ensemble of \acp{CNN}. To ensure high generalizability, the models were trained on a diverse dataset comprising both clinical trial and real-world data, spanning the full spectrum of the disease: from early-onset to late-stage \ac{GA}, as well as \ac{iAMD}, \ac{nAMD}, and healthy controls. The method has been validated on an independent real-world dataset. Furthermore, our evaluation was strengthened by adding (1)~a stratified analysis by lesion size, (2)~a reproducibility study, (3)~an inter-reader reliability assessment, and (4)~an analysis of the relevance of OCT B-Scan density on RPE loss and EZ loss computation, to assess the accuracy, reader agreement, and test--retest repeatability of fully automated biomarker quantification on OCT.

\section{Methods}
\subsection{Data acquisition}
All \ac{SD-OCT} scans have been acquired on a Spectralis device (Heidelberg Engineering GmbH, Heidelberg, Germany). 
The development dataset comprised \ac{SD-OCT} volumes sourced from three distinct cohorts. These included: (1)~the FILLY Phase-II clinical trial~\cite{liaoComplementC3Inhibitor2020} (\href{https://clinicaltrials.gov/study/NCT02503332}{NCT02503332}); (2)~a multi-device study from the Macula Clinic at the Medical University of Vienna (MUV), featuring patients with \ac{iAMD} and \ac{GA}~\cite{kostolnaSystematicProspectiveComparison2024}; and (3)~a subset of iAMD and GA cases from the Vienna Imaging Biomarker Eye Study (VIBES) real-world registry~\cite{gerendasVALIDATIONAUTOMATEDFLUID2022}. 
An additional independent validation dataset was assembled using separate scans from VIBES, ensuring that development and validation sets were mutually exclusive at the patient level. For the reproducibility study, data were utilized from the OAKS and DERBY Phase-III clinical trials~\cite{heierPegcetacoplanTreatmentGeographic2023} (\href{https://clinicaltrials.gov/study/NCT03525613}{NCT03525613} and \href{https://clinicaltrials.gov/study/NCT03525600}{NCT03525600}). 
For the B-scan density analysis, a \ac{SD-OCT} dataset with dense B-scan sampling (193 B-scans per volume) of patients with \ac{GA} has been incorporated~\cite{tratnig-franklAutomatedOCTtailoredBiomarker2025}.
All studies adhered to the Declaration of Helsinki, with informed consent obtained from all participants.
Scans were fovea-centered with patterns ranging from $49\times512$ to $97\times1024$ (B-scans $\times$ A-scans) in the development, validation and reproducibility datasets, and $193\times512$ in the B-scan density dataset, covering an approximate $6\times6 \si{\milli\metre}$ ($20\degree$) field of view. 

\subsection{Data preparation}
\paragraph {Data selection:}
To ensure model robustness, the development set (n=298 eyes) was curated for maximum phenotypic variability. This included a broad range of lesion sizes, edge cases challenging prior automated segmentation iterations~\cite{lachinovProjectiveSkipConnectionsSegmentation2021}, and non-GA controls. The latter encompassed \ac{iAMD} featuring \ac{iRORA}~\cite{saddaConsensusDefinitionAtrophy2018} and drusen, \ac{nAMD} with associated retinal fluid and subretinal fibrosis, and healthy eyes with normal aging phenotypes. 

The validation set comprised 43 eyes of 43 patients with \ac{GA} from the VIBES registry, selected in two steps by stratified random sampling. First, 19 scans were randomly sampled from VIBES, stratified by RPE loss area, for an internal validation of a previous algorithm version. Second, 24 additional scans were drawn from 950 VIBES visits with confirmed \ac{GA}, excluding patients already selected in the first step: one random visit per patient was selected, scans were grouped into quintiles of RPE loss area and drusen volume, and one scan was randomly drawn from each quintile combination. Drusen volume was quantified using a previously validated in-house segmentation model. The sample size was determined a priori from the measurement error of the previous algorithm version (paired comparison, $\alpha = 0.05$, power $= 0.9$), indicating that approximately 40 eyes were required.

The reproducibility dataset consisted of paired screening and baseline scans from the OAKS and DERBY trials. Selection was restricted to treatment-naïve eyes with an inter-visit interval of less than 10 days to avoid relevant GA progression~\cite{coulibalyProgressionDynamicsEarly2023a}. Furthermore, baseline scans had to be spatially registered to their respective screening volumes to ensure anatomical alignment.

\begin{figure}[tbp]
	\centering
	\includegraphics[width=0.7\linewidth]{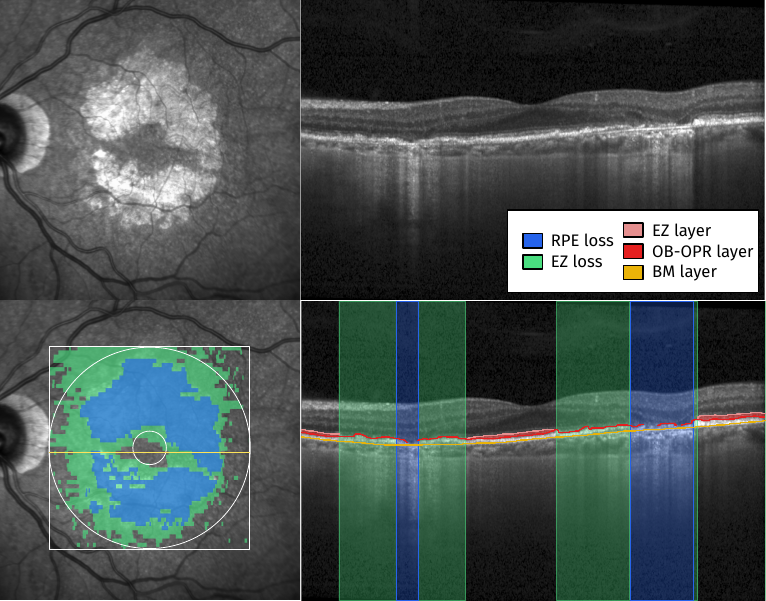}
	\caption{Annotation example. En-face view (left) and central B-scan (right) showing \acf{RPE} loss (blue) and \acf{EZ} loss (green) annotations, as well as the layer annotations of \acf{IB-EZ} in light red, \acf{OB-OPR} in dark red, and \acf{BM} in yellow.}
	\label{fig:annotation}
\end{figure}

\paragraph {Annotation of development and validation dataset:}
Ground truth was established by four experienced graders according to a clinically validated protocol. \ac{RPE} loss was defined as the complete absence of the \ac{EZ} band, downward displacement (subsidence) of the \ac{ONL} and \ac{OPL}, and substantial irregularities or absence of the typically continuous hyperreflective \ac{RPE} band. While choroidal hypertransmission was used to localize atrophic lesions, it was not used for the final delineation of boundaries. There was no minimum size restriction regarding the \ac{GA} loss area. 
Graders also delineated retinal and subretinal layers, precisely the \ac{BM}, \ac{IB-EZ}, and \ac{OB-OPR}.
The latter is commonly identical to the inner boundary of the \ac{RPE} layer, except in presence of subretinal fluid, or deposits such as \ac{SDD}, fibrotic tissue or blood. In that case \ac{OB-OPR} is delineated above these materials. \Ac{EZ} loss was identified by any interruption in the \ac{EZ} band, including focal disruptions caused by \ac{SDD} or pigment migration. A manual annotation example is provided in Fig. \ref{fig:annotation}.

\textit{Development set.} Annotators refined initial automated segmentations generated by previously described algorithms~\cite{lachinovProjectiveSkipConnectionsSegmentation2021, orlandoAutomatedQuantificationPhotoreceptor2020}. Each volume was annotated by a single grader and reviewed by the grading supervisor; difficult cases were additionally reviewed by a retinal specialist. RPE loss was annotated on every B-scan of 49-B-scan volumes and on every second B-scan of higher-density volumes; EZ layer boundaries were annotated on every fifth (49-B-scan volumes) or every tenth B-scan (higher-density volumes).

\textit{Validation set.} To avoid bias from automated pre-segmentations, the validation set was annotated de novo. Each volume was annotated by one of the four graders and subsequently reviewed and corrected by a second grader; a retinal specialist approved all final annotations. RPE loss and EZ loss were annotated on every B-scan of 49-B-scan volumes and on every second B-scan of 61- and 97-B-scan volumes, resulting in 49 annotated B-scans per volume (31 for 61-B-scan volumes). EZ layer boundaries were annotated separately on every fifth (49-B-scan volumes) or every tenth B-scan (61- and 97-B-scan volumes).

For both sets, ambiguous findings were discussed in weekly consensus meetings with the grading supervisor and a senior retinal specialist. All annotations were performed using pixel-accurate annotation software developed at the Medical University of Vienna.

To prevent data leakage, the validation set was not utilized for model training or hyperparameter tuning. Developers only gained access to the validation annotations after full completion of model development.

\subsection{Model training}
\begin{figure}[tbp]
	\centering
	\includegraphics[width=0.7\linewidth]{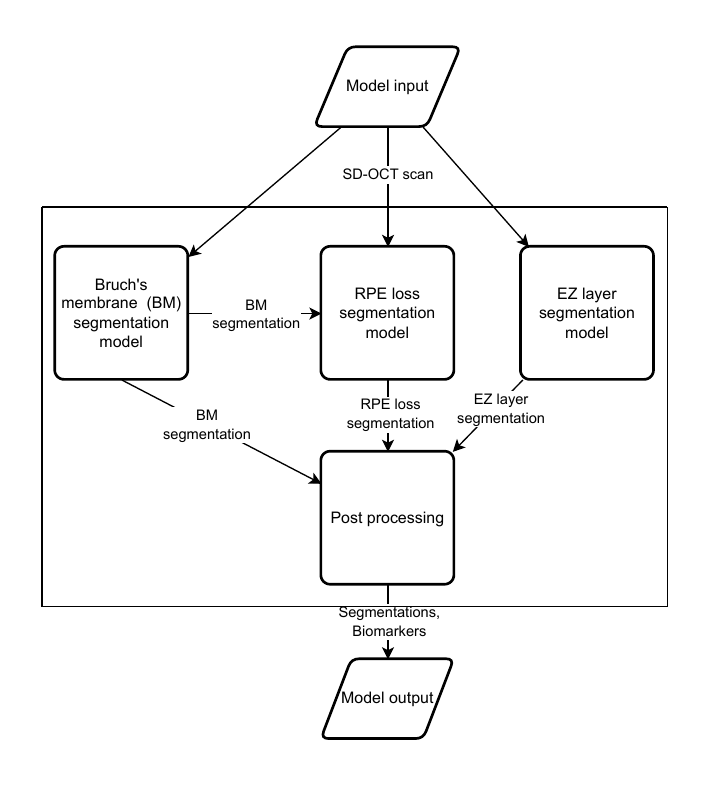}
	\caption{Data flow diagram describing the inputs and outputs of each model.}
	\label{fig:fig_1}
\end{figure}

Three semantic segmentation models were developed to segment \ac{RPE} loss, \ac{EZ} and \ac{BM} layer boundaries on SD-OCT scans (Fig. \ref{fig:fig_1}). BM segmentation was utilized to flatten the retinal curvature, providing a normalized input for the subsequent RPE loss model. During post-processing, outputs from all three models were integrated to generate the final segmentation and compute relevant biomarkers.
2D architectures were sourced from the PyTorch Image Models (TIMM) library~\cite{rw2019timm}, while the 3D-to-2D model was based on three candidate models evaluated by Morano et al.~\cite{moranoSelfsupervisedLearningIntermodal2023}.
Final architectures and hyperparameters were selected based on peak performance during validation set experiments.

\paragraph {\ac{BM} segmentation model}
\ac{BM} segmentation was implemented using the SD-Layernet framework~\cite{fazekasSegmentationBruchsMembrane2023, fazekasSDLayerNetSemisupervisedRetinal2022}, leveraging a topological engine to incorporate anatomical priors. We employed a 2D CNN UNet++~\cite{zhouUNetRedesigningSkip2020} featuring an EfficientNet-B5 encoder~\cite{tanEfficientNetRethinkingModel2019} to predict BM positions per individual A-scan at subpixel resolution. Training involved diverse stochastic augmentations: spatial shifts (zoom, left-right flip, translation, tilting, and bending), noise injection (speckle, Gaussian), intensity variations (linear and non-linear contrast changes, contrast gradients, artificial vessel shadows).
The model was optimized via a composite loss function comprising \ac{DSC} and \ac{CE} for layer masks, \ac{MAE} for layer positions, and topological constraints for curvature and continuity~\cite{fazekasSDLayerNetSemisupervisedRetinal2022, fazekasSegmentationBruchsMembrane2023}. Training was conducted for 80 epochs using the Adam optimizer (LR=$2\times10^{-4}$), with the final model selected based on the lowest validation \ac{MAE}. Development dataset splitting followed a 76:12:12 ratio (train/val/test), grouped by patient and stratified by RPE loss area and disease label. 

\paragraph {\ac{EZ} layer segmentation model}
The \ac{EZ} was segmented using a 2D U-Net architecture featuring a DenseNet-201 encoder~\cite{huangDenselyConnectedConvolutional2017}. Segmentation masks were defined across three distinct anatomical regions: from the \ac{IB-EZ} to the \ac{OB-OPR}, from the \ac{OB-OPR} to \ac{BM}, and from  \ac{BM} to the bottom of the B-scan. The model was trained for 80 epochs via the Adam optimizer (LR=$1.8\times10^{-4}$, weight decay = $3\times10^{-5}$). Final model selection was based on a weighted \ac{DSC} metric (0.3 for the \ac{EZ} region and 0.7 for \ac{EZ} interruptions). The dataset split and the stochastic augmentations remained identical to the BM segmentation.

\paragraph {\ac{RPE} loss segmentation model}
RPE loss was segmented using a 3D-to-2D \ac{FPN}~\cite{moranoSelfsupervisedLearningIntermodal2023}, which transforms a full 3D OCT volume into a 2D en-face segmentation map. To reduce spatial variance and standardize the input, each scan was first flattened relative to \ac{BM} using the segmentation output from the BM model. A-scans were vertically shifted to center the BM layer in the B-scan. Stochastic training augmentations were consistent with those used in other models, excluding tilt/bend transforms that are obsolete in flattened images, and the addition of random spatial cropping and B-scan order flipping. The model was trained using a balanced composite loss (0.5 \ac{DSC} and 0.5 \ac{CE}) optimized via \ac{SGD} (LR=$0.05$, momentum =$0.85$, weight decay =$3.5\times10^{-4}$). We implemented a "Reduce Learning Rate on Plateau" scheduler with a patience of 150 epochs, a 20-epoch cooldown, and a 0.5 reduction factor. Each model within the cross-validation framework was trained for 500 epochs using 16-bit mixed precision and a batch size of 16. For volumes with more than 49 B-scans, only every second B-scan was used as model input during training, with the B-scan spacing adjusted accordingly, so that all training inputs had an inter-B-scan spacing comparable to a 49-B-scan volume (approximately 120 \si{\micro\metre}). At inference, the full volume is processed at its native B-scan density, as the 3D-to-2D architecture is independent of the number of input B-scans. Model selection was based on the mean \ac{DSC} score for RPE loss. 
The development dataset was partitioned into training and hold-out test sets using the same partitions as for the EZ segmentation model. For model training, we merged the training and validation fold and utilized 5-fold cross-validation within the training set (4 folds for training, 1 for validation).

\paragraph {Post-processing}
During post-processing, the probability maps from the five cross-validation models for RPE loss were averaged to form an ensemble prediction. This ensemble was then binarized using a threshold of 0.5 to generate a final 2D en face RPE loss segmentation map. EZ loss maps were derived from the EZ layer segmentation by binarizing each A-scan based on the presence or absence of the EZ region.
To enhance segmentation robustness and ensure anatomical consistency, we integrated the model outputs using a heuristic constraint: since RPE loss in \ac{GA} regularly occurs within regions of \ac{EZ} loss \cite{leeSequentialStructuralFunctional2023, maiQuantitativeComparisonAutomated2024}, RPE loss predictions were dropped in A-scans where the corresponding EZ model did not indicate a loss.
All reported automated measurements were obtained without manual initialization or correction of the algorithm output.

\subsection{Evaluation and Statistical Analysis}
\paragraph{Accuracy of \ac{RPE}-loss and \ac{EZ}-loss area measurement}
The performance of RPE-loss and EZ-loss area measurements was evaluated against manual \ac{GT} annotations from the validation dataset. Total loss areas ($A$) were calculated by multiplying the cumulative pixel count of the lesion area by the physical voxel dimensions (B-scan and A-scan spacing). Measurement accuracy was quantified using \ac{AD} and \ac{PD}:
\begin{equation}
	  \label{eq:accuracy}
	\begin{aligned}
		\mathit{AD} &= |A_{\mathit{pred}} - A_{\mathit{gt}}|,\\
		\mathit{PD} &= \frac{AD}{A_\mathit{gt}} \times 100,
	\end{aligned}
\end{equation}
where $A_{\mathit{pred}}$ and $ A_{\mathit{gt}}$ are predicted and \ac{GT} loss areas, respectively. The model was applied to the full volumes, but all metrics were computed exclusively on annotated B-scans; no interpolation of the reference annotations was performed. For volumes annotated on every second B-scan, the B-scan spacing was doubled for area computation.
		
To assess the relationship between predicted and GT areas, Pearson’s R correlation was employed, while Bland-Altman~\cite{blandStatisticalMethodsAssessing1986a} analysis was used to identify potential proportional bias relative to lesion size. Additionally, \ac{DR}~\cite{linnetEvaluationRegressionProcedures1993} was performed to evaluate systematic bias, while accounting for potential measurement errors within the ground truth.
Pixel-level segmentation accuracy was assessed on en-face maps using following segmentation metrics based on the confusion matrix: \ac{DSC}, sensitivity, specificity, precision, \ac{NPV}, \ac{HD95}, and \ac{ASSD}~\cite{yeghiazaryanFamilyBoundaryOverlap2018}. 

All metrics were calculated per OCT volume and aggregated across the dataset using mean, \ac{SD}, median, and \ac{IQR}. For all mean estimates, 95\% \acp{CI} were determined via bias-corrected and accelerated (BCa) bootstrap resampling (1,000 iterations). The mean aggregation of \ac{AD} and \ac{PD} are equivalent to the common regression metrics of \ac{MAE} and \ac{MAPE}, respectively. For Bland-Altman plot \ac{LoA} the \acp{CI} were computed using an exact parametric method for paired \acp{LoA}~\cite{carkeetExactParametricConfidence2015}. 

Given that \ac{DSC} and \ac{PD} are inherently sensitive to total area size~\cite{seghierImageSegmentationEvaluation2024}, where minor segmentation errors disproportionately affect scores for smaller lesions, performance was further analysed across subgroups stratified by lesion size.

\paragraph{Accuracy of layer segmentations and \ac{EZ} thickness measurement:}
We evaluated the segmentation performance of the \ac{EZ} layers on the 43 eyes of the validation dataset. For each A-scan of the B-scans with layer annotations, manual \ac{GT} positions of the \ac{IB-EZ} and \ac{OB-OPR} were compared with the predicted positions. Performance was quantified using the \ac{MAE} of the individual layer positions and of the \ac{EZ} thickness (distance from the \ac{IB-EZ} to the \ac{OB-OPR}), averaged per volume over all A-scans in which both boundaries were present. Systematic bias of the thickness measurement was assessed by the signed mean difference and Bland-Altman \acp{LoA}.
 
\paragraph{Inter-reader reliability:}
To assess inter-reader reliability, two independent readers who were not involved in the initial labeling process annotated \ac{RPE} and \ac{EZ} loss on a subset of the validation dataset (annotation group AN2).
These readers were trained by the expert readers that provided the \ac{GT} (AN 1) according to the established protocol; however, all AN 2 annotations were performed without further supervision. To ensure a fair comparison with the iterative refinement process used by AN 1, the two AN 2 readers each annotated half of the subset and then cross-validated and corrected each other's work in a second iteration.

Inter-reader reliability was assessed using a two-way random-effects model to estimate the \ac{ICC} for single-score absolute agreement(ICC(2,1))~\cite{kooGuidelineSelectingReporting2016}. This model was chosen to account for both subject and reader variability, allowing for the generalization of results to other trained readers. We also determined the mean deviation by \ac{DR} and \acp{LoA} using Bland-Altman plots to compare the reader groups and the algorithm. Finally, segmentation accuracy was quantified using \ac{DSC}, \ac{HD95}, and \ac{ASSD} across all three segmentation sets.

\paragraph{Reproducibility:}
To evaluate reproducibility, we compared paired RPE and EZ loss areas between the screening and baseline visits of the reproducibility dataset. Agreement was assessed using ICC(2,1), \ac{LoA} (Bland-Altman plots), and mean deviation (\ac{DR}). To account for the hierarchical structure of the data, 95\% confidence intervals for the regression coefficients were estimated via nested bootstrap resampling with 1,000 iterations. The percentage difference between visits was computed relative to the mean of both measurements. To quantify the smallest change detectable in an individual eye, the repeatability coefficient (RC $= 2.77 \times$ within-subject \ac{SD}), corresponding to the minimal detectable change at 95\% confidence, was computed for RPE and EZ loss areas on the original and square-root-transformed scale, and for the mean \ac{EZ} thickness, with 95\% BCa bootstrap \acp{CI}.

Statistical analyses were performed using a combination of R and Python. The R package \textit{mcr} (v1.3.3.1) was utilized for \ac{DR} calculations, while the \textit{irr} package (v0.84.1) was used for \ac{ICC} estimation. All remaining statistical computations were conducted in Python 3.11 using the \textit{NumPy} (v1.26.4), \textit{SciPy} (v1.12.0), and \textit{Statsmodels} (v0.14.12) libraries.

\paragraph{Relevance of B-scan density on measurements:}
Spectralis OCT systems allow for variable B-scan spacing ranging from 11 to 300 µm, corresponding to 193 down to 19 B-scans in a 6 mm volumetric scan (standard defaults: 49 for high-speed; 97 for high-resolution). To determine the effect of B-scan density on the quantification accuracy of EZ and RPE loss, we first applied our segmentation algorithm to high-density scans (193 B-scans) to establish ground-truth reference area measurements. We then successively simulated lower B-scan densities by artificially increasing the inter-slice distances of the high-resolution segmentation maps via downsampling B-scans in the segmentation maps within a range of 192 to 19 B-scans. Biomarkers were recomputed at these simulated lower densities, and measurement variance was assessed against the ground truth using \ac{MAE} and \ac{MAPE}. Additionally, to assess the impact of scan density on measuring \ac{GA} progression, we analyzed a subset of cases with a 3-month follow-up. Progression was defined as the change in RPE and EZ loss area from baseline to month 3. Following the same protocol, \ac{GA} progression computed from the high-resolution scans served as the ground truth to evaluate the error in the lower-resolution simulations.

\section{Results}
\paragraph{Data sets and baseline characteristics:}
For model training and internal validation, 298 OCT volumes from 229 eyes of 221 patients were used. This cohort comprised 222 scans with \ac{GA}, 40 with \ac{iAMD} and drusen, 17 with \ac{nAMD}, and 19 healthy controls. The independent validation set consisted of 43 volumes from 43 eyes of 43 patients, all diagnosed with \ac{GA}. The validation set reflected typical real-world imaging conditions: at least one B-scan showed \acp{SDD} in 32 eyes (74\%), shadowing by hyperreflective foci in 17 (40\%), vitreous opacities in 16 (37\%), intraretinal fluid in 12 (28\%), out-of-registration artifacts in 8 (19\%), and increased noise in 5 (12\%). From this independent validation set, a subset of 25 volumes was selected for the inter-reader reliability analysis. Additionally, 68 \ac{OCT} volume pairs (screening and baseline visits) of 68 eyes from 68 patients in the OAKS and DERBY trials were included for the reproducibility study. 
Finally, for the B-scan density study, 61 baseline scans from 44 patients and 45 3-month follow-up scans (median interval 87 days, range 70--118) from 31 patients were included. 
Detailed baseline characteristics for all datasets are summarized in Table \ref{tab:baseline_characteristics}.

\begin{table}[tbp]
	
	\caption{Population characteristics for the datasets used for training, independent validation, as well as reproducibility, inter-reader reliability, and B-scan density analyses. Lesion size is reported as mean ± std | median ± IQR.}
	\label{tab:baseline_characteristics}
	\begin{footnotesize}
		\setlength{\tabcolsep}{2pt}
		\begin{tabular}{@{}L{.24\textwidth}L{.15\textwidth}L{.14\textwidth}L{.14\textwidth}L{.14\textwidth}L{.14\textwidth}@{}}
			\toprule
			\textbf{Dataset (Source)} &
			\textbf{Training} &
			\textbf{Independent validation\newline (VIBES)} &
			\textbf{Repro\-ducibility\newline (OAKS \& DERBY)} &
			\textbf{Inter-reader reliability validation\newline (VIBES)} & \textbf{B-scan density high-resolution scans} 
			\\ \midrule
			\textbf{Number of patients / eyes / OCTs} & 221 / 229 / 298 & 43 / 43 / 43 & 68 / 68 / 136 & 25 / 25 / 25 & 44 / 61 / 106 \\
			\textbf{\# annotated B-scans}             & RPE loss: 14602 |\newline EZ~layer: 2910 & RPE/EZ loss: 2071 |\newline EZ~layer: 434 & N/A           & 1189  & N/A        \\
			\textbf{Age mean ± std}                   & N/A & 78.44 ± 8.12 & 77.72 ± 5.83  & N/A  & 79.1 ± 5.03         \\
			\textbf{Gender female \%}                 & 63 & 55           & 71            & N/A   & 59        \\
			\textbf{RPE loss area {[}mm²{]}} &
			6.16 ± 5.25 |\newline 5.46 ± 8.40  & 5.30 ± 7.27 |\newline 1.86 ± 6.74 & 7.73 ± 3.40 |\newline 7.15 ± 5.17 &  3.64 ± 4.47 |\newline 1.87 ± 4.41 & 5.38 ± 4.23 |\newline 4.02 ± 7.24 \\
			\textbf{EZ loss area {[}mm²{]}} & 
			N/A & 11.32 ± 10.57 |\newline 7.19 ± 14.58 & 14.50 ± 6.63 |\newline 13.36 ± 8.25 & 10.91 ± 10.40 |\newline 7.59 ± 12.09 & 10.08 ± 5.93 |\newline 8.91 ± 9.71 \\ \bottomrule
		\end{tabular}
	\end{footnotesize}
\end{table}

\begin{figure}[tbp]
	\centering
	\includegraphics[width=\linewidth]{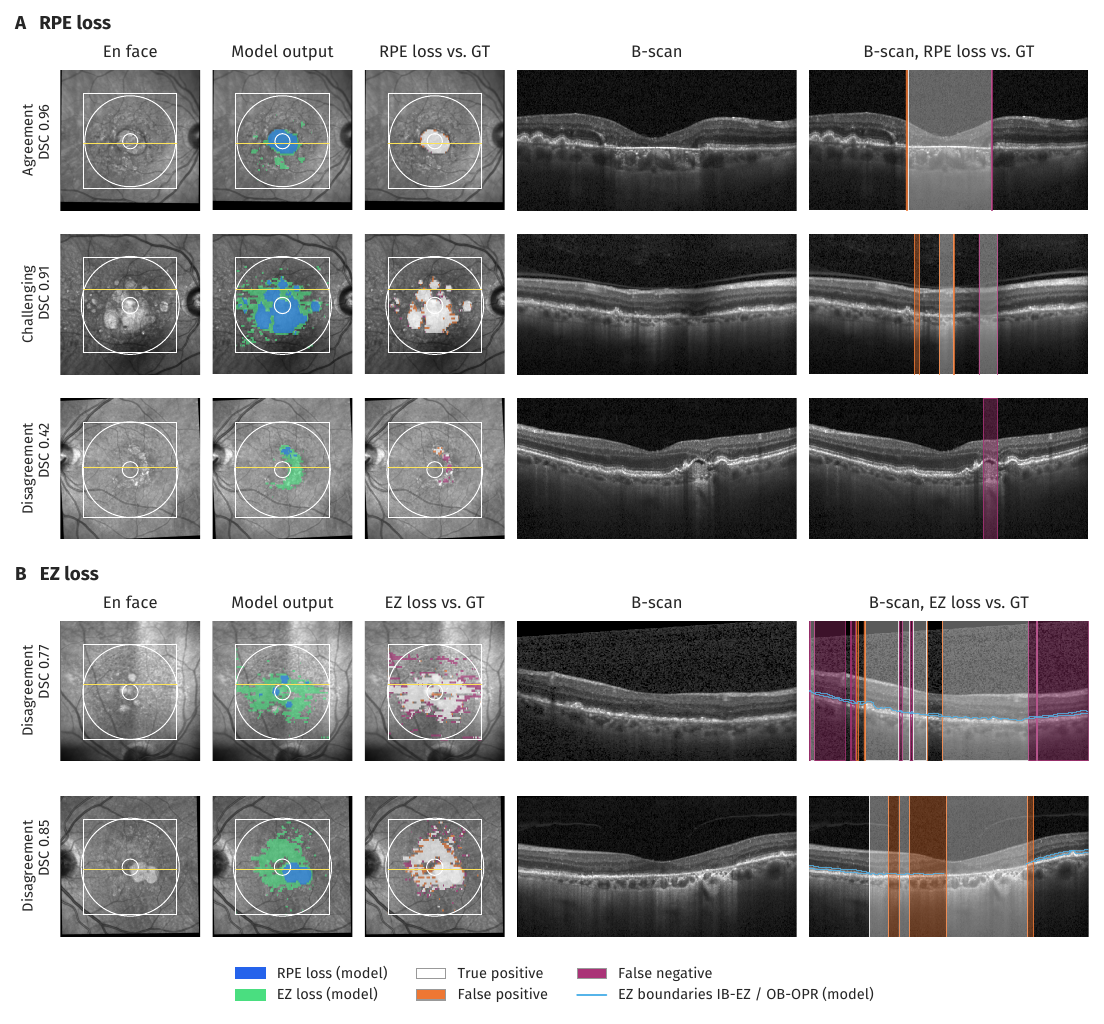}
	\caption{Representative automated segmentations of (A) \ac{RPE} loss and (B) \ac{EZ} loss from the independent validation dataset. (A) Agreement with the manual annotation; a challenging multifocal lesion with small satellite lesions; and disagreement in early atrophy with a partially preserved RPE band, which the readers annotated as RPE loss while the model did not. (B) Disagreement in a scan with low signal, where the model still detected a faint EZ band that the readers no longer considered present (under-segmentation relative to \ac{GT}); and disagreement where the model merged adjacent atrophic areas across residual EZ that the readers still considered present (over-segmentation). Columns: infrared en-face image with B-scan position (yellow line); model output (RPE loss blue, EZ loss green); agreement with \ac{GT} (true positive white, false positive orange, false negative purple); B-scan; B-scan with agreement and, in (B), predicted \ac{IB-EZ} and \ac{OB-OPR} boundaries (light blue).}
	\label{fig:examples}
\end{figure}

\paragraph{Accuracy of \ac{RPE}-loss and \ac{EZ}-loss area measurement:}
Performance metrics are summarized in Table \ref{tab:performance_area_measurement}, with their distributions visualized via boxplots in Fig. \ref{fig:boxplot}. Detailed subset analyses of the stratified data are provided in the Supplemental Material (Table S1, Figure S1). Representative examples of agreement, a challenging case and disagreement between automated and manual segmentation are shown in Fig. \ref{fig:examples}. Overall, the model achieved a mean \ac{DSC} of 0.88 (95\% CI: 0.84 to 0.90) for RPE loss and 0.87 (95\% CI: 0.83 to 0.89) for EZ loss. Stratification by lesion size revealed the correlation between loss area and overall DSC; performance for RPE loss ranged from 0.84 (0.5--2.5 \si{\milli\metre\squared}) to 0.96 ($>10$ \si{\milli\metre\squared}), while EZ loss \ac{DSC} ranged from 0.76 ($\leq 2.5$ \si{\milli\metre\squared}) to 0.95 ($>15$ \si{\milli\metre\squared}).

\begin{figure}[tbp]
	\centering
	\includegraphics[width=0.7\linewidth]{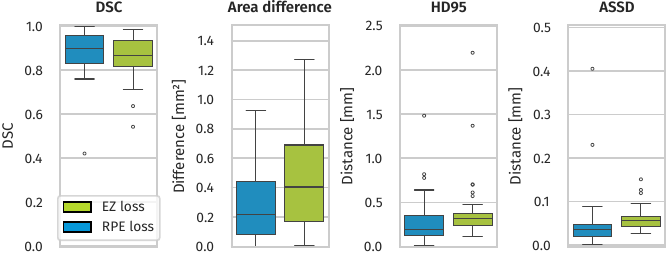}
	\caption{Distribution of segmentation evaluation metrics comparing manual and automated segmentations of \ac{RPE} loss (blue) and \ac{EZ} loss (green) in the independent validation dataset. Reported metrics are the absolute difference between measured and annotated lesion size, \acf{DSC} for segmentation overlap, \acf{HD95} 95\% and \acf{ASSD} for lesion surface accuracy.}
\label{fig:boxplot}
\end{figure}

\begin{table}[tbp]
	\caption{Comparison of the total RPE-loss and EZ-loss area measurements between manual ground truth annotation and automated segmentation. Numbers in brackets are the 95\% \acf{CI}.}
	\label{tab:performance_area_measurement}
	\begin{footnotesize}
		\setlength{\tabcolsep}{2pt}
		\begin{tabular}{@{}L{.195\textwidth}L{.41\textwidth}L{.37\textwidth}@{}}
			\toprule
			\textbf{Metric} & \textbf{RPE loss}                 & \textbf{EZ loss} \\ \midrule 	
			\textbf{Pearson’s r}                           & 0.999 {[}0.998 to 1.000{]}       & 0.996 {[}0.993 to 0.998{]}      \\
			\textbf{\ac{DR} Intercept {[}mm²{]}} & 0.14 {[}0.047 to 0.28{]}          & -0.56 {[}-0.93 to -0.27{]}       \\
			\textbf{\ac{DR} Slope{[}mm²{]}}      & 1.03 {[}1.015 to 1.05{]}          & 1.02 {[}0.99 to 1.05{]}          \\ \cmidrule(l){2-3}					
			& \multicolumn{2}{c}{\textbf{Mean {[}95\% CI{]} ± std | median ± IQR}} \\ \cmidrule(l){2-3}
			\textbf{\ac{AD} loss area {[}mm²{]}}           & 0.34 {[}0.24 to 0.50{]} ± 0.40 | 0.22 ± 0.36       & 0.63 {[}0.46 to 0.93{]} ± 0.76 | 0.40 ± 0.52  \\
			\textbf{\ac{PD} loss area {[}\%{]}} & 14.24 {[}10.60 to 18.53{]} ± 13.31 | 10.18 ± 15.54 & 8.57 {[}6.46 to 11.87{]} ± 8.73 | 6.39 ± 7.35 \\ 
			\textbf{\ac{DSC}}                     & 0.88 {[}0.84 to 0.90{]} ± 0.10 | 0.90 ± 0.12 & 0.87 {[}0.83 to 0.89{]} ± 0.10 | 0.87 ± 0.12 \\
			\textbf{Sensitivity}             & 0.92 {[}0.87 to 0.95{]} ± 0.12 | 0.96 ± 0.06 & 0.84 {[}0.80 to 0.87{]} ± 0.12 | 0.87 ± 0.17 \\
			\textbf{Specificity}             & 0.98 {[}0.97 to 0.99{]} ± 0.03 | 0.99 ± 0.02 & 0.96 {[}0.93 to 0.97{]} ± 0.07 | 0.98 ± 0.03 \\
			\textbf{Precision}               & 0.84 {[}0.81 to 0.88{]} ± 0.10 | 0.87 ± 0.15 & 0.90 {[}0.87 to 0.92{]} ± 0.08 | 0.92 ± 0.08 \\
			\textbf{\ac{NPV}} & 0.99 {[}0.99 to 1.00{]} ± 0.01 | 1.00 ± 0.01 & 0.94 {[}0.92 to 0.96{]} ± 0.06 | 0.97 ± 0.06 \\ 
			\textbf{\ac{HD95} {[}mm{]}}           & 0.29 {[}0.23 to 0.41{]} ± 0.27 | 0.20 ± 0.23 & 0.39 {[}0.32 to 0.56{]} ± 0.35 | 0.32 ± 0.13 \\
			\textbf{\ac{ASSD} {[}mm{]}}           & 0.05 {[}0.04 to 0.08{]} ± 0.07 | 0.04 ± 0.03 & 0.06 {[}0.05 to 0.07{]} ± 0.03 | 0.06 ± 0.02 \\ \bottomrule			
		\end{tabular}
	\end{footnotesize}\\
	\scriptsize{\Acf{AD}, \acf{PD}, \acf{DR}, \acf{AD}, \acf{PD}, \acf{DSC}, \acf{NPV}, \acf{HD95}, \acf{ASSD}}
\end{table}

\begin{figure}[tbp]
	\centering	
	\includegraphics[width=0.8\linewidth]{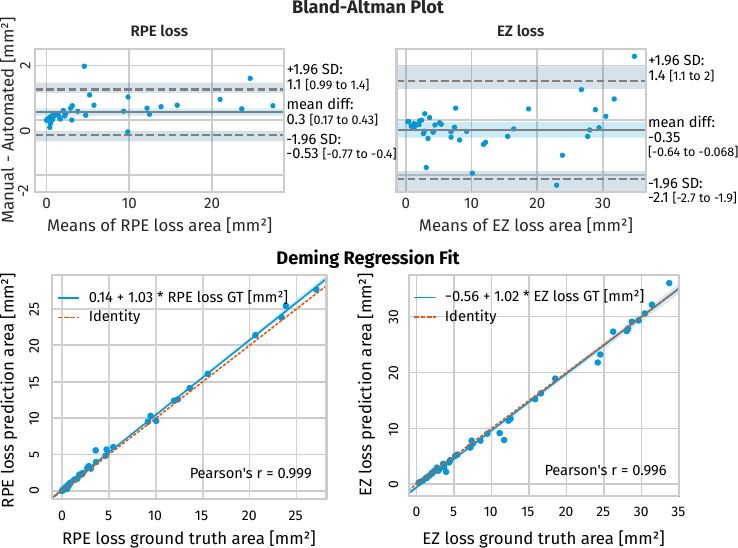}
	\caption{Bland Altman plot (top) and Deming regression fit (bottom) from automated segmentation compared to manual ground truth annotation for RPE loss areas (left) and EZ loss areas (right). Numbers in brackets are the 95\% \acf{CI}.}
	\label{fig:blandaltmandeming}
\end{figure}

Area measurements demonstrated high correlation between manual and automated segmentations, with Pearson’s r of 0.999 (95\% CI: 0.998 to 1.0) for RPE loss and 0.996 (95\% CI: 0.993 to 0.998) for EZ loss. Deming regression parameters and Bland-Altman analysis (Fig. \ref{fig:blandaltmandeming}) indicated a slight systematic bias: a marginal over-segmentation of the RPE loss area (mean difference: 0.3 \si{\milli\metre\squared} (95\% CI: 0.17 to 0.43)) and a marginal under-segmentation of the EZ loss area (mean difference: -0.35 \si{\milli\metre\squared} (95\% CI: -0.64 to -0.068)).

The anatomical constraint affected 20 of 43 validation eyes and removed a mean RPE loss area of 0.006 \si{\milli\metre\squared} (maximum 0.118 \si{\milli\metre\squared}), corresponding to 0.12\% of all predicted RPE loss pixels; its effect on the reported metrics is therefore negligible.

\paragraph{Accuracy of layer segmentation and \ac{EZ} thickness measurement:}
As summarized in Table \ref{tab:performance_layer_position}, the mean differences between manual and predicted positions for the \ac{IB-EZ}, \ac{OB-OPR}, and total \ac{EZ} thickness were 2.78 \si{\micro\metre}, 4.32 \si{\micro\metre}, and 2.08 \si{\micro\metre}, respectively. Given a pixel spacing of 3.9 \si{\micro\metre}, the average deviations for both \ac{IB-EZ} position and \ac{EZ} thickness are less than a single pixel, demonstrating high accuracy in an OCT setting. The automated \ac{EZ} thickness was on average 1.24 \si{\micro\metre} lower than the manual reference (95\% CI: -1.77 to -0.52 \si{\micro\metre}; Bland-Altman \acp{LoA}: -5.42 to 2.94 \si{\micro\metre}), corresponding to -4.4\% of the mean annotated thickness of 28.10 \si{\micro\metre}.

\begin{table}[tbp]
	\caption{Layer segmentation performance. \Ac{AD} between ground truth position of \ac{IB-EZ} and \ac{OB-OPR}, as well as \ac{AD} and \ac{PD} for EZ thickness are reported.}
	\label{tab:performance_layer_position}
	\begin{center}
	\begin{footnotesize}
		\begin{tabular}{@{}L{.37\textwidth}L{.37\textwidth}@{}}
			\toprule
			\textbf{Metric}                                           & \textbf{Mean {[}95\% CI{]} ± std | median ± IQR} \\ \midrule
			\textbf{AD IB-EZ layer position} [\si{\micro\metre}]  & 2.78 {[}2.57 to 3.09{]} ± 0.86 | 2.51 ± 0.90                         \\
			\textbf{AD OB-OPR layer position} [\si{\micro\metre}] & 4.32 {[}4.07 to 4.64{]} ± 0.95 | 4.20 ± 1.18                         \\
			\textbf{AD IB-EZ to OB-OPR thickness} [\si{\micro\metre}] & 2.08 {[}1.72 to 2.49{]} ± 1.30 | 1.92 ± 1.91                     \\
			\textbf{PD IB-EZ to OB-OPR thickness} [\%]                & 7.26 {[}6.04 to 8.56{]} ± 4.26 | 6.93 ± 6.65 \\
			\textbf{Signed difference thickness (prediction $-$ GT)} [\si{\micro\metre}] & -1.24 {[}-1.77 to -0.52{]} ± 2.13 | -1.67 ± 2.67 \\
            \textbf{DSC IB-EZ to OB-OPR} & 0.83 {[}0.81 to 0.85{]} ± 0.07 | 0.85 ± 0.08 \\ \bottomrule			
		\end{tabular}
	\end{footnotesize}
	\end{center}
	\scriptsize{\Acf{AD}, \acf{PD}, \acf{IB-EZ}, \acf{OB-OPR}, \acf{DSC}}
\end{table}

\begin{figure}[tbp]
	\centering	
	\includegraphics[width=0.8\linewidth]{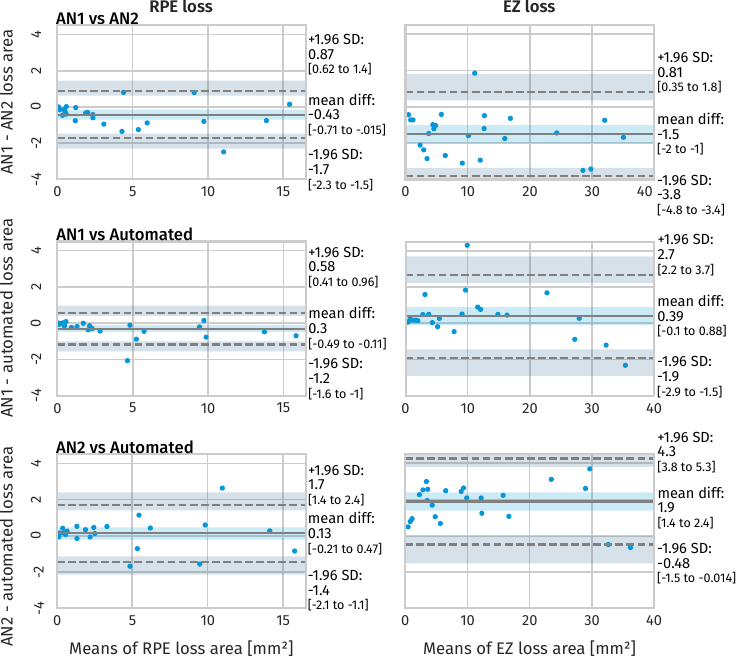}
	\caption{Bland Altman plots for RPE loss area(left) and EZ loss area (right) inter-reader reliability comparing reader groups AN1 vs AN2 (top), AN1 vs automated segmentation (center) and AN2 vs automated segmentation (bottom). Numbers in brackets are the 95\% \acf{CI}.}
	\label{fig:irr_bland_altman}
\end{figure}

\paragraph{Inter-reader reliability:}
Table \ref{tab:interreader_reliability} presents the inter-reader reliability between the two annotation groups (AN1 and AN2). The ICC(2,1) values for RPE and EZ loss area measurements were 0.985 (95\% CI: 0.948 to 0.994) and 0.984 (95\% CI: 0.743 to 0.996), respectively. Bland-Altman analysis (Fig. \ref{fig:irr_bland_altman}) and Deming regression intercepts (Table \ref{tab:interreader_reliability}, Supplemental Figure S2) revealed a systematic bias in EZ loss measurements, with AN2 consistently over-segmenting the area relative to AN1. For mean RPE and EZ loss lesion sizes of 3.65 \si{\milli\meter\squared} and 10.91 \si{\milli\meter\squared}, the \acp{LoA} were -1.7 to 0.87 \si{\milli\meter\squared} and -3.8 to 0.81 \si{\milli\meter\squared}, respectively. Notably, the automated segmentation showed closer alignment with AN1, the group responsible for the training dataset.
Segmentation overlap between the two reader groups (\ac{DSC}: 0.76 for RPE loss and 0.79 for EZ loss; \ac{HD95}: 0.55 and 0.57 \si{\milli\metre}) was lower than between AN1 and the algorithm (\ac{DSC}: 0.88 and 0.84; \ac{HD95}: 0.27 and 0.53 \si{\milli\metre}), and similar to that between AN2 and the algorithm (\ac{DSC}: 0.78 and 0.77; \ac{HD95}: 0.53 and 0.60 \si{\milli\metre}).

\begin{table}[tbp]
	\caption{Inter-reader reliability in terms of \ac{ICC}, \ac{DR} coefficients and Pearson’s R statistics for agreement between annotator groups AN 1, AN 2 and automated segmentation. }
	\label{tab:interreader_reliability}
	\begin{center}
	\begin{footnotesize}
	\setlength{\tabcolsep}{2pt}
	\begin{tabular}{@{}L{.08\textwidth}L{.19\textwidth}L{.195\textwidth}L{.10\textwidth}L{.19\textwidth}L{.19\textwidth}@{}}
		\toprule
		&        & \textbf{ICC(2,1)}            & \textbf{Pearsons R} & \textbf{\ac{DR} Intercept}     & \textbf{\ac{DR} Slope}        \\ \midrule
		\textbf{RPE loss} & \textbf{AN2 vs AN1} & 0.985 {[}0.948 to 0.994{]} & 0.990 & 0.3 {[}0.06 to 0.51{]} & 1.0 {[}0.97 to 1.16{]} \\
		& \textbf{AN1 vs automated} & 0.993 {[}0.975 to 0.997{]} & 0.996               & 0.16 {[}0.045 to 0.34{]}  & 1.04 {[}1.012 to 1.08{]} \\
		& \textbf{AN2 vs automated} & 0.985 {[}0.966 to 0.993{]} & 0.984               & -0.14 {[}-0.36 to 0.11{]} & 1.00 {[}0.89 to 1.09{]}  \\ \midrule
		\textbf{EZ loss}  & \textbf{AN2 vs AN1} & 0.984 {[}0.743 to 0.996{]}  & 0.994 & 1.2 {[}0.60 to 1.8{]}  & 1.0 {[}0.98 to 1.1{]}  \\
		& \textbf{AN1 vs automated} & 0.993 {[}0.985 to 0.997{]}  & 0.994               & -0.8 {[}-1.38 to -0.29{]} & 1.0 {[}0.97 to 1.07{]}   \\
		& \textbf{AN2 vs automated} & 0.978 {[}0.493 to 0.995{]} & 0.994               & -2.1 {[}-2.80 to -1.2{]}  & 1.0 {[}0.93 to 1.1{]}    \\ \bottomrule
	\end{tabular}
	\end{footnotesize}
	\end{center}
	\scriptsize{\Acf{ICC}, \acf{DR}}
\end{table}

\begin{figure}[tbp]
	\centering	
	\includegraphics[width=0.8\linewidth]{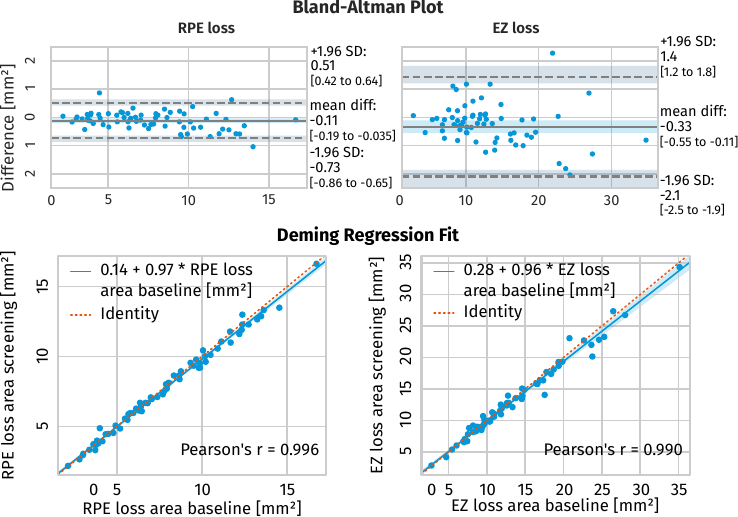}
	\caption{Reproducibility study: Bland Altman plot (top) and Deming regression fit (bottom) from automated segmentation of screening and baseline visit for RPE loss areas (left) and EZ loss areas (right). Numbers in brackets are the 95\% \acf{CI}.}
	\label{fig:reproducibility_blandaltmandeming}
\end{figure}

\paragraph{Reproducibility:}
The comparison of automated segmentations between screening and baseline visits from the OAKS and DERBY trials is summarized in Table \ref{tab:reproducibility}. Reproducibility was excellent, with ICC(2,1) values of 0.995 (95\% CI: 0.991 to 0.997) for RPE loss and 0.988 (95\% CI: 0.978 to 0.993) for EZ loss. Deming regression slopes were close to unity (RPE loss: 0.97, 95\% CI: 0.943 to 0.99; EZ loss: 0.96, 95\% CI: 0.91 to 1.00; Fig. \ref{fig:reproducibility_blandaltmandeming}), with slightly larger areas at the baseline than at the screening visit (mean difference: 0.11 \si{\milli\metre\squared} for RPE loss and 0.33 \si{\milli\metre\squared} for EZ loss; visits 3 to 9 days apart). The mean percentage difference in lesion area between the two visits was 3.33\% for RPE loss and 4.70\% for EZ loss, demonstrating high reproducibility of the automated measurements. The repeatability coefficient, i.e., the minimal detectable change between two measurements of the same eye, was 0.66 \si{\milli\metre\squared} (95\% CI: 0.54 to 0.82) for RPE loss and 1.88 \si{\milli\metre\squared} (95\% CI: 1.46 to 2.55) for EZ loss, or 0.11 \si{\milli\metre} (95\% CI: 0.09 to 0.15) and 0.23 \si{\milli\metre} (95\% CI: 0.19 to 0.31) on the square-root scale. For the mean \ac{EZ} thickness, the ICC(2,1) was 0.968 (95\% CI: 0.941 to 0.981), the mean absolute difference $0.68 \pm 0.55$ \si{\micro\metre} (2.47\%), and the repeatability coefficient 1.72 \si{\micro\metre} (95\% CI: 1.45 to 2.03).

\begin{table}[tbp]
	\caption{Comparing automated segmentations in the reproducibility study from screening visit and baseline visit in terms of RPE and EZ loss areas. Numbers in brackets are the 95\% \acf{CI}.}
	\label{tab:reproducibility}
	\begin{center}
	\begin{footnotesize}
	\begin{tabular}{@{}L{.24\textwidth}L{.27\textwidth}L{.27\textwidth}@{}}
		\toprule
		& \textbf{RPE loss}          & \textbf{EZ loss}           \\ \midrule
		& \multicolumn{2}{c}{\textbf{Mean ± std | median ± IQR}}  \\ \cmidrule(l){2-3} 
		\textbf{Area at baseline} [\si{\milli\metre\squared}]& 7.764 ± 3.389 | 7.543 ± 5.190 & 13.472 ± 6.228 | 11.874 ± 7.721 \\
		\textbf{\ac{AD} loss area} [\si{\milli\metre\squared}]    & 0.251 ± 0.223 | 0.160 ± 0.268 & 0.644 ± 0.718 | 0.468 ± 0.746    \\
		\textbf{\ac{PD} loss area} [\%]& 3.33 ± 2.97 | 2.81 ± 2.97  & 4.70 ± 4.28 | 3.40 ± 4.93  \\ \midrule
		\textbf{\ac{ICC}(2,1)}     & 0.995 {[}0.991 to 0.997{]} & 0.988 {[}0.978 to 0.993{]} \\
		\textbf{Pearson's R}  & 0.996 {[}0.994 to 0.998{]} & 0.990 {[}0.984 to 0.994{]} \\
		\textbf{\ac{DR} Intercept} & 0.14 {[}-0.022 to 0.32{]}  & 0.28 {[}-0.14 to 0.79{]}   \\
		\textbf{\ac{DR} Slope}     & 0.97 {[}0.943 to 0.99{]}   & 0.96 {[}0.91 to 1.00{]}   \\
		\textbf{Repeatability coefficient} [\si{\milli\metre\squared}] & 0.66 {[}0.54 to 0.82{]} & 1.88 {[}1.46 to 2.55{]} \\
		\textbf{Repeatability coefficient, square-root scale} [\si{\milli\metre}] & 0.11 {[}0.09 to 0.15{]} & 0.23 {[}0.19 to 0.31{]} \\ \bottomrule
	\end{tabular}
	\end{footnotesize}
	\end{center}
	\scriptsize{\Acf{AD}, \acf{PD}, \acf{ICC}, \acf{DR}, \acf{IQR}}
\end{table}

\begin{figure}[tbp]
	\centering	
	\includegraphics[width=.8\linewidth]{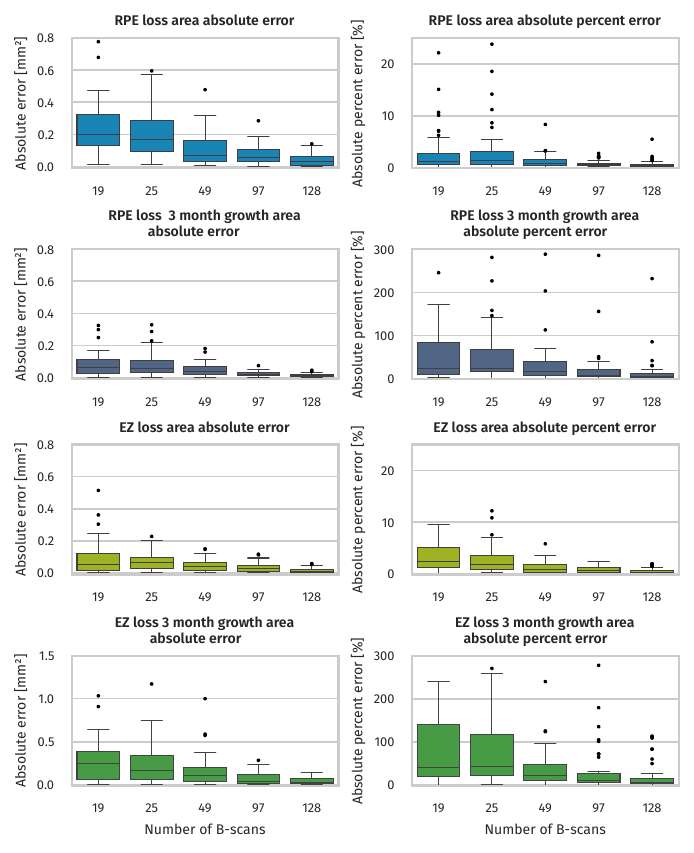}
	\caption{Effect of B-scan density on RPE loss and EZ loss measurement. Boxplot showing distribution of absolute errors (left) and absolute percentage errors (right) of RPE loss area, RPE loss progression area, EZ loss areas, and EZ loss progression area (top to bottom), comparing synthetically downsampled segmentations in the range of 19 to 128 B-Scans to ground-truth segmentation with 193 B-scans.}
	\label{fig:b_scan_density}
\end{figure}

\paragraph{B-scan density:}
Table \ref{tab:bscan_density} reports the \ac{MAE} and \ac{MAPE} for RPE loss, EZ loss, and their progression areas across standard B-scan patterns: 19, 25, 49, and 97 (Spectralis scanners) and 128 (Zeiss Cirrus and Topcon Maestro2 scanners), compared to a 193 B-scan baseline. Corresponding B-scan spacings were 314, 239, 122, 61, 47, and 31 \si{\micro\meter} . Error distributions are shown in Fig. \ref{fig:b_scan_density}, and Supplemental Figure S3 lists errors for all B-scan spacings from 19 to 192.

\begin{sidewaystable}[tbp]
\caption{Effect of B-scan density on RPE loss and EZ loss measurement. Loss area and its change within 3 months is compared for \ac{GT} segmentation using 193 B-scans with synthetical downsampled segmentations simulating reduced B-scan density in the range of 19 to 128 B-Scans. \acf{MAE},\acf{MAPE} and their corresponding \acf{SD} are reported. Numbers in brackets are the 95\% \acf{CI}.}
\label{tab:bscan_density}
\begin{center}
\begin{footnotesize}
\begin{tabular}{@{}C{.07\textwidth}L{.16\textwidth}L{.16\textwidth}L{.16\textwidth}L{.20\textwidth}@{}}
\hline
                  & \multicolumn{2}{l}{\textbf{GT loss area{[}mm²{]} (n=61)}}       & \multicolumn{2}{l}{\textbf{GT 3 month  progression absolute loss area {[}mm²{]} (n=45)}} \\ \cline{2-5} 
\textbf{RPE loss} & \multicolumn{2}{l}{5.38 ± 4.23 | 4.02 ± 7.24}                               & \multicolumn{2}{l}{0.32 ± 0.29 | 0.19 ± 0.40}                                         \\
\textbf{EZ loss}  & \multicolumn{2}{l}{10.08 ± 5.93 | 8.91 ± 9.71}                                & \multicolumn{2}{l}{0.61 ± 0.55 | 0.51 ± 0.70}                                         \\ \hline
\textbf{\# B-scans} & \textbf{\ac{MAE} loss area  {[}mm²{]}} & \textbf{\ac{MAPE} loss area {[}\%{]}} & \textbf{\ac{MAE} progression \newline loss area  {[}mm²{]}} & \textbf{\ac{MAPE} progression \newline loss area {[}\%{]}} \\ \hline
\textbf{RPE loss} & \textbf{}                      &                                &                                 &                                         \\
\textbf{19}       & 0.08 {[}0.06 to 0.11{]} ± 0.10 & 2.52 {[}1.77 to 3.88{]} ± 3.83 & 0.08 {[}0.06 to 0.11{]} ± 0.07  & 106.71 {[}58.29 to 193.06{]} ± 228.82   \\
\textbf{25}       & 0.07 {[}0.06 to 0.09{]} ± 0.06 & 2.81 {[}2.01 to 4.37{]} ± 4.32 & 0.08 {[}0.06 to 0.11{]} ± 0.08  & 95.80 {[}52.11 to 176.62{]} ± 200.22    \\
\textbf{49}       & 0.04 {[}0.04 to 0.05{]} ± 0.04 & 1.20 {[}0.95 to 1.68{]} ± 1.28 & 0.05 {[}0.04 to 0.07{]} ± 0.04  & 93.16 {[}40.22 to 243.58{]} ± 288.92    \\
\textbf{97}       & 0.03 {[}0.03 to 0.04{]} ± 0.03 & 0.78 {[}0.66 to 0.93{]} ± 0.55 & 0.02 {[}0.02 to 0.03{]} ± 0.02  & 29.66 {[}16.43 to 64.46{]} ± 68.65      \\
\textbf{128}      & 0.02 {[}0.01 to 0.02{]} ± 0.01 & 0.54 {[}0.41 to 0.82{]} ± 0.80 & 0.01 {[}0.01 to 0.02{]} ± 0.01  & 23.09 {[}9.78 to 53.13{]} ± 66.48       \\ \hline
\textbf{EZ loss}  & \multicolumn{4}{l}{\textbf{}}                                                                                                               \\
\textbf{19}       & 0.24 {[}0.20 to 0.28{]} ± 0.16 & 3.21 {[}2.61 to 3.91{]} ± 2.53 & 0.28 {[}0.21 to 0.35{]} ± 0.24  & 222.97 {[}126.79 to 395.48{]} ± 438.00  \\
\textbf{25}       & 0.21 {[}0.17 to 0.24{]} ± 0.15 & 2.68 {[}2.19 to 3.45{]} ± 2.50 & 0.25 {[}0.19 to 0.33{]} ± 0.25  & 93.13 {[}63.21 to 145.06{]} ± 130.72    \\
\textbf{49}       & 0.10 {[}0.08 to 0.13{]} ± 0.10 & 1.27 {[}1.00 to 1.56{]} ± 1.11 & 0.15 {[}0.11 to 0.22{]} ± 0.18  & 88.60 {[}52.12 to 152.89{]} ± 163.84    \\
\textbf{97}       & 0.07 {[}0.06 to 0.09{]} ± 0.06 & 0.83 {[}0.69 to 0.99{]} ± 0.63 & 0.07 {[}0.05 to 0.09{]} ± 0.07  & 32.14 {[}20.59 to 56.67{]} ± 53.34      \\
\textbf{128}      & 0.04 {[}0.03 to 0.05{]} ± 0.03 & 0.50 {[}0.38 to 0.63{]} ± 0.53 & 0.05 {[}0.04 to 0.06{]} ± 0.04  & 20.05 {[}12.77 to 31.50{]} ± 31.28      \\ \hline
\end{tabular}
\end{footnotesize}
\end{center}
\scriptsize{\Acf{MAE}, \acf{MAPE}, \acf{GT}}
\end{sidewaystable}

GA lesion progression was slow during the 3-month interval, making the progression area small relative to overall lesion size (e.g., 0.32 vs. 5.38 \si{\milli\meter\squared} for RPE). Combined with large B-scan spacings, this resulted in high percentage errors, such as a 95\% error for RPE loss progression using 25 B-scans (a 100\% error equates to a factor-of-two overestimation). The primary cause of these errors is that the physical lesion progression is smaller than the spacing between slices, rendering the changes unobservable.

\section {Discussion}
Based on the work performed in the described path, we are able to present a deep learning-based algorithm for high-quality, high accuracy and fully automated segmentation of pivotal \ac{GA} features in conventional retinal \ac{OCT} images applicable to real-world settings and spanning a wide range of morphological conditions. The focus of our work are the two hallmark features relevant for \ac{GA} monitoring in clinical trials and routine, i.e. \ac{RPE} loss and outer photoreceptor degeneration (represented by \ac{EZ} thinning and loss). We evaluated the accuracy of automated area and thickness measurements, as well as the precision of the segmentations in patients with a large range of GA manifestations secondary to AMD. Overall, we report high agreement between \acf{GT} and automated segmentations, with a mean \ac{DSC} of 0.88 and 0.87, and a Pearson’s r of 0.999 and 0.996 for RPE and EZ loss area, respectively.

\paragraph{Accuracy of RPE and EZ loss segmentation and measurements:} 
Area measurements demonstrated strong correlation between automated and manual segmentations. For very large lesions ($>  20$ \si{\milli\metre\squared}), the algorithm tends to report slightly higher values, a trend visualized in the Deming regression and Bland-Altman plots. The high specificity (RPE loss: 0.98, EZ loss: 0.96) indicates that non-atrophic areas were rarely misclassified as atrophic. 
A high sensitivity and slightly lower precision for RPE loss (0.92 and 0.84, respectively) suggests a small trend towards over-segmentation. Qualitative analysis revealed that such "false positives" typically occur at irregular lesion borders or in regions showing early signs of atrophy that have not yet met the strict reader criteria for "complete" RPE loss, but which already manifest signs of atrophy. For a clinical assessment of RPE integrity, the borders are biologically not as distinct as jumping from completely healthy to completely lost RPE. The anatomical constraint thus acts as a safeguard against isolated false-positive RPE loss detections rather than contributing substantially to the reported segmentation performance.
In EZ segmentation sensitivity is marginally lower than precision (0.84 and 0.90), indicating a trend to slightly under-segment EZ loss compared to \ac{GT}. As the \ac{EZ} layer is primarily a thin and less reflective band, remaining \ac{EZ} fragments at the zone of highest disease activity and cellular debris are expected. Both findings highlight the limitations of advanced image analysis technology by biology and visualization. Qualitative review of the eyes with the lowest agreement revealed three recurring patterns: early atrophy with a partially preserved RPE band, which readers annotated as RPE loss while the algorithm did not; barely visible EZ bands, for which both readers and algorithm were inconsistent; and small focal EZ interruptions caused by \acp{SDD} (Fig. \ref{fig:examples}).

\paragraph{Subgroup analysis of RPE and EZ loss:} 
Training and validation datasets frequently exclude very small or very large lesions, potentially introducing selection bias.~\cite{monesRateProgressionGeographic2018}. 
To reduce this risk, we included real-world data covering the full spectrum of GA progression, from early-stage to advanced lesions. Our stratified analysis further underscores an important limitation of \ac{DSC} and other confusion-matrix–based metrics, including sensitivity and specificity, as these measures are strongly influenced by lesion size~\cite{seghierImageSegmentationEvaluation2024}. For small lesions ($\leq 0.5$\si{\milli\metre\squared}), even minor boundary errors can disproportionately affect performance estimates relative to larger lesions.
When considering common clinical trial size restrictions (2.5 to 17.5 \si{\milli\metre\squared} in FILLY, OAKS and DERBY), our DSC increased to 0.91 (95\% CI: 0.87 to 0.93) for RPE loss and 0.93 (95\% CI: 0.92 to 0.94) for EZ loss (n = 15).
In contrast, surface distance metrics such as \ac{HD95} and \ac{ASSD} proved more robust to size variations. We found that 95\% of all automated segmentation surface points were within $0.29 \pm 0.27$ \si{\milli\metre} (RPE) and $0.39 \pm 0.35$ \si{\milli\metre} (EZ) of the manual annotations. \ac{HD95} and \ac{ASSD} in subgroups stayed in the same range, independent on lesion size. 
The detection of small lesions is clinically relevant, as the biology of growth differs from larger ones \cite{coulibalyProgressionDynamicsEarly2023a}.

\paragraph{Comparison with State-of-the-art:}
Despite the sensitivity of metrics to lesion size, our results compare favourably to current methods. For RPE loss, our DSC of 0.88 is entirely consistent with Lachinov et al.~\cite{lachinovProjectiveSkipConnectionsSegmentation2021} (0.92), al-Khersan et al.\cite{al-khersanDeepLearningBasedSegmentation2025} (0.82), Spaide et al.~\cite{spaideEstimatingUncertaintyGeographic2025} (0.90), and the method by Morano et al.~\cite{moranoDeepMultimodalFusion2024} (0.89) upon which our segmentation method is based. Yoshida et al. \cite{yoshidaDeepLearningApproaches2025} report a Pearson $R^2$ of $0.91$ (our method: $0.998$).

For EZ loss, our DSC of 0.87 outperforms Pfau et al.~\cite{pfauProgressionPhotoreceptorDegeneration2020} (0.82), and our layer position \ac{MAE} of $2.78 \pm 0.86$ \si{\micro\metre} is even lower than the $3.92 \pm 3.72$ \si{\micro\metre} reported by Mishra et al.~\cite{mishraAutomatedRetinalLayer2020}. 
Riedl et al.~\cite{riedlEffectPegcetacoplanTreatment2022} report a mean \ac{DSC} of $0.838 \pm 0.084$ for automatic segmentation of the \ac{EZ} to \ac{OB-OPR} region. In comparison, we report a comparable mean DSC of $0.83 \pm 0.07$.
However, comparison with other methods is limited due to different datasets evaluated and different annotation criteria used.

Precise and reliable determination of \ac{EZ} boundaries is critical. Although complete \ac{EZ} loss is often considered a point of morphological no-return, microperimetry demonstrates that retinal sensitivity in these areas ranges between 8 and 10 dB, indicating substantial residual activity of "invisible" photoreceptors \cite{birnerStructureFunctionCorrelationDeepLearning2025, ansariEvaluatingProgressionRetinal2025a}. Structure-function analyses indicate that EZ attenuation is a continuous rather than binary process; for instance, partial EZ loss represents a pathway of slowly progressive thinning~\cite{kalraAutomatedIdentificationSegmentation2023}. Clinically, baseline EZ thinning is a major risk factor for both the conversion from \ac{iAMD} to \ac{GA}~\cite{voglSpatiotemporalAlterationsRetinal2021a} and the subsequent progression rate of manifest lesions~\cite{voglPredictingTopographicDisease2023}. Localized EZ thickness accurately predicts the risk of conversion at specific topographical points along the GA margin~\cite{schmidt-erfurthLongitudinalAssessmentProgressive2025}. Furthermore, the ratio between EZ loss area and RPE loss area has been shown as an predictor for future GA progression rate~\cite{maiDynamicsEZRPE2025}. However, the reported quartiles of ratios based on the OAKS \& DERBY study eyes are slightly higher using this segmentation algorithm (compare Tables from \cite{maiDynamicsEZRPE2025} with Supplemental Table S2) due to higher sensitivity in detecting EZ loss. Crucially, EZ thickness serves as a strong structural predictor of retinal function, exhibiting a well-defined correlation with microperimetry sensitivity (µm/dB)~\cite{birnerStructureFunctionCorrelationDeepLearning2025}. Moving forward, the integration of novel high-resolution imaging devices, such as the High-Resolution Spectralis \cite{frank-publigQuantificationsOuterRetinal2025}, will further enhance the precision of EZ segmentation.

\paragraph{Inter-reader reliability:}
The inter-reader study confirms that RPE loss can be annotated reliably across different groups (ICC 0.985), with the algorithm successfully reproducing these results. Reader-versus-algorithm ICC aligns closely with inter-reader ICC, indicating that the algorithm performs on par with human annotators while producing identical results for identical input. EZ segmentation remains more challenging~\cite{leeChallengesAssociatedEllipsoid2021}, requiring strict and detailed annotation protocols and expert training. Although inter-group agreement was high (ICC 0.984), we observed a slight bias: the second reader group (AN2) tended to classify "barely visible" EZ bands as loss, whereas the first group (AN1) classified them as present. Interestingly, the algorithm aligned more closely with AN1 for both loss types, who provided also the initial training data. Importantly, agreement between the algorithm and the independent reader group AN2 was in the same range as agreement between the two reader groups, indicating that the deviations of the algorithm lie within human inter-reader variability rather than merely reproducing the annotation style of the training graders.

However, the reproducibility study validated that RPE loss and EZ loss can be segmented consistently, showing a high agreement between screening and baseline visits (ICC:~0.988; mean absolute percentage difference:~4.70\%). On the square-root scale, the repeatability coefficient of RPE loss (0.11 \si{\milli\metre}) corresponds to approximately four months of untreated lesion growth, given a mean square-root growth of 0.35 \si{\milli\metre} per year in the sham arm of the FILLY trial~\cite{liaoComplementC3Inhibitor2020}, although growth in that trial was measured on fundus autofluorescence.

\paragraph{Effect of B-scan density on RPE and EZ loss measurement:}

Simulating lower B-scan densities showed that total lesion area can be measured accurately even with sparse patterns (e.g., a mean absolute percentage error of 1.2\% for RPE loss with 49 B-scans). The 3-month change in lesion area, however, was much more sensitive to B-scan spacing: with 49 or fewer B-scans, its error was of the same order as the change itself (mean absolute percentage error of approximately 90\% or more), decreasing to approximately 30\% with 97 B-scans and 20 to 23\% with 128 B-scans. As \ac{GA} progresses slowly, with an approximate progression rate of 1 mm per year~\cite{voglPredictingTopographicDisease2023}, and with high local variability, dense patterns of at least 97 B-scans are recommended when short-term change is of interest.

\paragraph{Limitations:}
A primary limitation is the B-scan density of the validation annotations: RPE and EZ loss were annotated on 49 B-scans per volume (31 for two 61-B-scan volumes), corresponding to a B-scan spacing of approximately 120 \si{\micro\metre} (approximately 200 \si{\micro\metre} for the 61-B-scan volumes). A single-slice discrepancy therefore corresponds to an error of approximately 120 \si{\micro\metre}, which particularly impacts distance metrics at lesion borders. 
Additionally, slight misalignments in follow-up scans in the reproducibility study can affect the segmentation of focal EZ interruptions caused by \acp{SDD}. However, as these interruptions are small relative to the total GA area, they did not significantly impact the overall reproducibility scores.

\paragraph{Conclusion:}
We developed a fully automated method for segmenting RPE loss, EZ loss, and EZ thickness on OCT. On independent real-world data, its accuracy was comparable to that of expert readers, its deviations from the reference annotation were within human inter-reader variability, and its measurements showed high test--retest repeatability. This makes it a candidate for standardized and scalable quantification of \ac{GA} in clinical trials, research, and routine ophthalmic care.

\section*{Acknowledgments}
Presented in part at the Annual Meeting of the Association for Research in Vision and Ophthalmology (ARVO), Salt Lake City, UT, USA, May 4--8, 2025.

Supported by Apellis, a subsidiary of Biogen, Inc., Cambridge, MA, USA and RetInSight GmbH, Vienna, Austria.

The segmentation algorithm, its source code, the model development, and the initial draft of this manuscript were created by the authors without the use of generative artificial intelligence (AI). During the revision of the manuscript, the authors used Gemini 3 Pro (Google, Mountain View, CA, USA; February 2026) and Claude Code (version 2.1.322; model Claude Opus 5.5; Anthropic, San Francisco, CA, USA; November 2026) to edit and revise the text and check reported values. The structure, scientific content, and interpretation of the manuscript are the authors' own. The authors reviewed and verified all AI-assisted content and take full responsibility for the content of the publication.

Disclosure: \textbf{W.-D. Vogl}, RetInSight (E); \textbf{H. Skulason}, RetInSight (E); \textbf{O. Leingang}, RetInSight (E); \textbf{U. Schmidt-Erfurth}, AbbVie (C), ADARx (C), Alcon (C), Alkeus (C), Apellis (C), Astellas (C), Aviceda (C), Bayer (C), Complement Therapeutics (C), Genentech (C), Kodiak (C), Medscape (C), Roche (C), Samsung (C), Topcon (C); \textbf{A. Sadeghipour}, RetInSight (E); \textbf{A. Whitby}, RetInSight (E)

\paragraph{Data availability.} Data underlying the results presented in this paper are not publicly available at this time but may be obtained from the authors upon reasonable request.

\clearpage

\bibliographystyle{myvancouver3}
\bibliography{references/biblio} %

\begin{thebibliography}{10}

\bibitem{chakravarthyCharacterizingDiseaseBurden2018}
Chakravarthy U, Bailey CC, Johnston RL, et~al.
\newblock Characterizing {{Disease Burden}} and {{Progression}} of {{Geographic
  Atrophy Secondary}} to {{Age-Related Macular Degeneration}}.
\newblock {\itshape Ophthalmology}. 2018;125(6):842--849.

\bibitem{holzGeographicAtrophyClinical2014}
Holz FG, Strauss EC, {Schmitz-Valckenberg} S, {van Lookeren Campagne} M.
\newblock Geographic {{Atrophy}}: {{Clinical Features}} and {{Potential
  Therapeutic Approaches}}.
\newblock {\itshape Ophthalmology}. 2014;121(5):1079--1091.

\bibitem{reiterAIClinicalManagement2024}
Reiter GS, Mai J, Riedl S, et~al.
\newblock {{AI}} in the Clinical Management of {{GA}}: {{A}} Novel Therapeutic
  Universe Requires Novel Tools.
\newblock {\itshape Progress in Retinal and Eye Research}. 2024;103:101305.

\bibitem{saddaConsensusDefinitionAtrophy2018}
Sadda SR, Guymer R, Holz FG, et~al.
\newblock Consensus {{Definition}} for {{Atrophy Associated}} with
  {{Age-Related Macular Degeneration}} on {{OCT}}: {{Classification}} of
  {{Atrophy Report}} 3.
\newblock {\itshape Ophthalmology}. 2018;125(4):537--548.

\bibitem{zekavatPhotoreceptorLayerThinning2022}
Zekavat SM, Sekimitsu S, Ye Y, et~al.
\newblock Photoreceptor {{Layer Thinning Is}} an {{Early Biomarker}} for
  {{Age-Related Macular Degeneration}}: {{Epidemiologic}} and {{Genetic
  Evidence}} from {{UK Biobank OCT Data}}.
\newblock {\itshape Ophthalmology}. 2022;129(6):694--707.

\bibitem{staurenghiProposedLexiconAnatomic2014}
Staurenghi G, Sadda S, Chakravarthy U, Spaide RF.
\newblock Proposed Lexicon for Anatomic Landmarks in Normal Posterior Segment
  Spectral-Domain Optical Coherence Tomography: The {IN}{\textbullet}{OCT}
  Consensus.
\newblock {\itshape Ophthalmology}. 2014;121(8):1572--1578.

\bibitem{reiterSubretinalDrusenoidDeposits2020}
Reiter GS, Told R, Schranz M, et~al.
\newblock Subretinal {{Drusenoid Deposits}} and {{Photoreceptor Loss Detecting
  Global}} and {{Local Progression}} of {{Geographic Atrophy}} by {{SD-OCT
  Imaging}}.
\newblock {\itshape Investigative Ophthalmology \& Visual Science}.
  2020;61(6):11.

\bibitem{pfauProgressionPhotoreceptorDegeneration2020}
Pfau M, {von der Emde} L, {de Sisternes} L, et~al.
\newblock Progression of {{Photoreceptor Degeneration}} in {{Geographic Atrophy
  Secondary}} to {{Age-related Macular Degeneration}}.
\newblock {\itshape JAMA Ophthalmology}. 2020;138(10):1026--1034.

\bibitem{riedlEffectPegcetacoplanTreatment2022}
Riedl S, Vogl WD, Mai J, et~al.
\newblock The {{Effect}} of {{Pegcetacoplan Treatment}} on {{Photoreceptor
  Maintenance}} in {{Geographic Atrophy Monitored}} by {{Artificial
  Intelligence}}--{{Based OCT Analysis}}.
\newblock {\itshape Ophthalmology Retina}. 2022;6(11):1009--1018.

\bibitem{voglPredictingTopographicDisease2023}
Vogl WD, Riedl S, Mai J, et~al.
\newblock Predicting {{Topographic Disease Progression}} and {{Treatment
  Response}} of {{Pegcetacoplan}} in {{Geographic Atrophy Quantified}} by
  {{Deep Learning}}.
\newblock {\itshape Ophthalmology Retina}. 2023;7(1):4--13.

\bibitem{maresCorrelationRetinalFluid2025}
Mares V, Reiter GS, Gumpinger M, et~al.
\newblock Correlation of Retinal Fluid and Photoreceptor and {{RPE}} Loss in
  Neovascular {{AMD}} by Automated Quantification, a Real-World {{FRB}}!
  Analysis.
\newblock {\itshape Acta Ophthalmologica}. 2025;103(3):295--303.

\bibitem{birnerStructureFunctionCorrelationDeepLearning2025}
Birner K, Reiter GS, Steiner I, et~al.
\newblock Structure-{{Function Correlation}} of {{Deep-Learning Quantified
  Ellipsoid Zone}} and {{Retinal Pigment Epithelium Loss}} and
  {{Microperimetry}} in {{Geographic Atrophy}}.
\newblock {\itshape Investigative Ophthalmology \& Visual Science}.
  2025;66(3):26.

\bibitem{birnerNormativeProspectiveData2025}
Birner K, Coulibaly LM, Schrittwieser J, et~al.
\newblock Normative Prospective Data on Automatically Quantified Retinal
  Morphology Correlated to Retinal Function in Healthy Ageing Eyes by Two
  Microperimetry Devices.
\newblock {\itshape Acta Ophthalmologica}. 2025;103(4):423--431.

\bibitem{maiDynamicsEZRPE2025}
Mai J, Reiter GS, Riedl S, et~al.
\newblock Dynamics of the {{EZ}}/{{RPE Loss Ratio}} on {{OCT Over Time During
  Geographic Atrophy Progression}} and {{Treatment With Pegcetacoplan}}.
\newblock {\itshape Investigative Ophthalmology \& Visual Science}.
  2025;66(11):48.

\bibitem{stealthbiotherapeuticsinc.ReNEWPhase32025}
{Stealth BioTherapeutics Inc }.
\newblock {{ReNEW}}: {{A Phase}} 3, {{Double-Masked}}, {{Placebo-Controlled
  Clinical Trial}} to {{Evaluate}} the {{Efficacy}}, {{Safety}}, and
  {{Pharmacokinetics}} of {{Subcutaneous Injections}} of {{Elamipretide}} in
  {{Subjects Who Have Dry Age-Related Macular Degeneration}} ({{Dry AMD}}).
\newblock clinicaltrials.gov; 2025. NCT06373731.

\bibitem{ruiz-morenoAutomaticQuantificationSoftware2020}
{Ruiz-Moreno} JM, {Ruiz-Medrano} J, Lugo F, Sirvent B, {Flores-Moreno} I.
\newblock Automatic {{Quantification Software}} for {{Geographic Atrophy
  Associated}} with {{Age-Related Macular Degeneration}}: {{A Validation
  Study}}.
\newblock {\itshape Journal of Ophthalmology}. 2020;2020(1):8204641.

\bibitem{lachinovProjectiveSkipConnectionsSegmentation2021}
Lachinov D, Seeb{\"o}ck P, Mai J, Goldbach F, {Schmidt-Erfurth} U, Bogunovic H.
\newblock Projective {{Skip-Connections}} for {{Segmentation Along}} a
  {{Subset}} of {{Dimensions}} in {{Retinal OCT}}.
\newblock In: De~Bruijne M, Cattin PC, Cotin S, et~al., editors. Medical
  {{Image Computing}} and {{Computer Assisted Intervention}} -- {{MICCAI}}
  2021. vol. 12901. Cham: Springer International Publishing; 2021:431--441.

\bibitem{pramilDeepLearningModel2023}
Pramil V, {de Sisternes} L, Omlor L, et~al.
\newblock A {{Deep Learning Model}} for {{Automated Segmentation}} of
  {{Geographic Atrophy Imaged Using Swept-Source OCT}}.
\newblock {\itshape Ophthalmology Retina}. 2023;7(2):127--141.

\bibitem{moranoDeepMultimodalFusion2024}
Morano J, Aresta G, Grechenig C, {Schmidt-Erfurth} U, Bogunovi{\'c} H.
\newblock Deep {{Multimodal Fusion}} of {{Data With Heterogeneous
  Dimensionality}} via {{Projective Networks}}.
\newblock {\itshape IEEE Journal of Biomedical and Health Informatics}.
  2024;28(4):2235--2246.

\bibitem{spaideEstimatingUncertaintyGeographic2025}
Spaide T, Rajesh AE, Gim N, et~al.
\newblock Estimating {{Uncertainty}} of {{Geographic Atrophy Segmentations}}
  with {{Bayesian Deep Learning}}.
\newblock {\itshape Ophthalmology Science}. 2025;5(1).

\bibitem{al-khersanDeepLearningBasedSegmentation2025}
{Al-khersan} H, Sodhi SK, Cao JA, et~al.
\newblock Deep {{Learning-Based Segmentation}} of {{Geographic Atrophy}}: {{A
  Multi-Center}}, {{Multi-Device Validation}} in a {{Real-World Clinical
  Cohort}}.
\newblock {\itshape Diagnostics}. 2025;15(20):2580.

\bibitem{orlandoAutomatedQuantificationPhotoreceptor2020}
Orlando JI, Gerendas BS, Riedl S, et~al.
\newblock Automated {{Quantification}} of {{Photoreceptor}} Alteration in
  Macular Disease Using {{Optical Coherence Tomography}} and {{Deep Learning}}.
\newblock {\itshape Scientific Reports}. 2020;10(1):5619.

\bibitem{kalraAutomatedIdentificationSegmentation2023}
Kalra G, Cetin H, Whitney J, et~al.
\newblock Automated {{Identification}} and {{Segmentation}} of {{Ellipsoid Zone
  At-Risk Using Deep Learning}} on {{SD-OCT}} for {{Predicting Progression}} in
  {{Dry AMD}}.
\newblock {\itshape Diagnostics}. 2023;13(6).

\bibitem{schmidt-erfurthDiseaseActivityTherapeutic2025}
{Schmidt-Erfurth} U, Mai J, Reiter GS, et~al.
\newblock Disease {{Activity}} and {{Therapeutic Response}} to
  {{Pegcetacoplan}} for {{Geographic Atrophy Identified}} by {{Deep
  Learning-Based Analysis}} of {{OCT}}.
\newblock {\itshape Ophthalmology}. 2025;132(2):181--193.

\bibitem{birnerExploringTrialEndpoints2026}
Birner K, Mai J, Boryshchuk D, et~al.
\newblock Exploring {{Trial Endpoints}} in {{Geographic Atrophy Based}} on
  {{Localized Functional Changes}} in {{Microperimetry}} and {{AI-Quantified
  OCT Biomarkers}}.
\newblock {\itshape Investigative Ophthalmology \& Visual Science}.
  2026;67(1):22.

\bibitem{suNavigatingDistributionShifts2025}
Su Z, Guo J, Yang X, Wang Q, Coenen F, Huang K. Navigating {{Distribution
  Shifts}} in {{Medical Image Analysis}}: {{A Survey}}.
\newblock arXiv; 2025.

\bibitem{liaoComplementC3Inhibitor2020}
Liao DS, Grossi FV, El~Mehdi D, et~al.
\newblock Complement {{C3 Inhibitor Pegcetacoplan}} for {{Geographic Atrophy
  Secondary}} to {{Age-Related Macular Degeneration}}: {{A Randomized Phase}} 2
  {{Trial}}.
\newblock {\itshape Ophthalmology}. 2020;127(2):186--195.

\bibitem{kostolnaSystematicProspectiveComparison2024}
Kostolna K, Reiter GS, Frank S, et~al.
\newblock A {{Systematic Prospective Comparison}} of {{Fluid Volume
  Evaluation}} across {{OCT Devices Used}} in {{Clinical Practice}}.
\newblock {\itshape Ophthalmology Science}. 2024;4(3).

\bibitem{gerendasVALIDATIONAUTOMATEDFLUID2022}
Gerendas BS, Sadeghipour A, Michl M, et~al.
\newblock Validation of an {{Automated Fluid Algorithm}} on {{Real-world Data}}
  of {{Neovascular Age-related Macular Degeneration Over Five Years}}.
\newblock {\itshape RETINA}. 2022;42(9):1673.

\bibitem{heierPegcetacoplanTreatmentGeographic2023}
Heier JS, Lad EM, Holz FG, et~al.
\newblock Pegcetacoplan for the Treatment of Geographic Atrophy Secondary to
  Age-Related Macular Degeneration ({{OAKS}} and {{DERBY}}): Two Multicentre,
  Randomised, Double-Masked, Sham-Controlled, Phase 3 Trials.
\newblock {\itshape The Lancet}. 2023;402(10411):1434--1448.

\bibitem{tratnig-franklAutomatedOCTtailoredBiomarker2025}
{Tratnig-Frankl} M, Kuchernig L, Birner K, et~al.
\newblock Automated {{OCT-tailored}} and Biomarker Targeted Microperimetry in
  Geographic Atrophy.
\newblock {\itshape Investigative Ophthalmology \& Visual Science}.
  2025;66(8):4787--4787.

\bibitem{coulibalyProgressionDynamicsEarly2023a}
Coulibaly LM, Reiter GS, Fuchs P, et~al.
\newblock Progression {{Dynamics}} of {{Early}} versus {{Later Stage Atrophic
  Lesions}} in {{Nonneovascular Age-Related Macular Degeneration Using
  Quantitative OCT Biomarker Segmentation}}.
\newblock {\itshape Ophthalmology Retina}. 2023;7(9):762--770.

\bibitem{rw2019timm}
Wightman R. {{PyTorch}} Image Models.
\newblock GitHub; 2019.

\bibitem{moranoSelfsupervisedLearningIntermodal2023}
Morano J, Aresta G, Lachinov D, Mai J, {Schmidt-Erfurth} U, Bogunovi{\'c} H.
\newblock Self-Supervised {{Learning}} via~{{Inter-modal Reconstruction}}
  and~{{Feature Projection Networks}} for~{{Label-Efficient 3D-to-2D
  Segmentation}}.
\newblock In: Greenspan H, Madabhushi A, Mousavi P, et~al., editors. Medical
  {{Image Computing}} and {{Computer Assisted Intervention}} -- {{MICCAI}}
  2023. Cham: Springer Nature Switzerland; 2023:589--599.

\bibitem{fazekasSegmentationBruchsMembrane2023}
Fazekas B, Lachinov D, Aresta G, Mai J, {Schmidt-Erfurth} U, Bogunovi{\'c} H.
\newblock Segmentation of {{Bruch}}'s {{Membrane}} in {{Retinal OCT With AMD
  Using Anatomical Priors}} and {{Uncertainty Quantification}}.
\newblock {\itshape IEEE Journal of Biomedical and Health Informatics}.
  2023;27(1):41--52.

\bibitem{fazekasSDLayerNetSemisupervisedRetinal2022}
Fazekas B, Aresta G, Lachinov D, et~al.
\newblock {{SD-LayerNet}}: {{Semi-supervised Retinal Layer Segmentation}}
  in~{{OCT Using Disentangled Representation}} with~{{Anatomical Priors}}.
\newblock In: Wang L, Dou Q, Fletcher PT, Speidel S, Li S, editors. Medical
  {{Image Computing}} and {{Computer Assisted Intervention}} -- {{MICCAI}}
  2022. Cham: Springer Nature Switzerland; 2022:320--329.

\bibitem{zhouUNetRedesigningSkip2020}
Zhou Z, Siddiquee MMR, Tajbakhsh N, Liang J.
\newblock {{UNet}}++: {{Redesigning Skip Connections}} to {{Exploit Multiscale
  Features}} in {{Image Segmentation}}.
\newblock {\itshape IEEE Transactions on Medical Imaging}.
  2020;39(6):1856--1867.

\bibitem{tanEfficientNetRethinkingModel2019}
Tan M, Le Q.
\newblock {{EfficientNet}}: {{Rethinking Model Scaling}} for {{Convolutional
  Neural Networks}}.
\newblock In: Proceedings of the 36th {{International Conference}} on {{Machine
  Learning}}. PMLR; 2019:6105--6114.

\bibitem{huangDenselyConnectedConvolutional2017}
Huang G, Liu Z, Van Der~Maaten L, Weinberger KQ.
\newblock Densely {{Connected Convolutional Networks}}.
\newblock In: 2017 {{IEEE Conference}} on {{Computer Vision}} and {{Pattern
  Recognition}} ({{CVPR}}); 2017:2261--2269.

\bibitem{leeSequentialStructuralFunctional2023}
Lee JH, Ahn J, Shin JY.
\newblock Sequential Structural and Functional Change in Geographic Atrophy on
  Multimodal Imaging in Non-Exudative Age-Related Macular Degeneration.
\newblock {\itshape Graefe's Archive for Clinical and Experimental
  Ophthalmology}. 2023;261(8):2199--2207.

\bibitem{maiQuantitativeComparisonAutomated2024}
Mai J, Reiter GS, Riedl S, et~al.
\newblock Quantitative Comparison of Automated {{OCT}} and Conventional
  {{FAF-based}} Geographic Atrophy Measurements in the Phase 3
  {{OAKS}}/{{DERBY}} Trials.
\newblock {\itshape Scientific Reports}. 2024;14(1):20531.

\bibitem{blandStatisticalMethodsAssessing1986a}
Bland JM, Altman DG.
\newblock Statistical Methods for Assessing Agreement between Two Methods of
  Clinical Measurement.
\newblock {\itshape Lancet (London, England)}. 1986;1(8476):307--310.

\bibitem{linnetEvaluationRegressionProcedures1993}
Linnet K.
\newblock Evaluation of Regression Procedures for Methods Comparison Studies.
\newblock {\itshape Clinical Chemistry}. 1993;39(3):424--432.

\bibitem{yeghiazaryanFamilyBoundaryOverlap2018}
Yeghiazaryan V, Voiculescu I.
\newblock Family of Boundary Overlap Metrics for the Evaluation of Medical
  Image Segmentation.
\newblock {\itshape Journal of Medical Imaging}. 2018;5(1):015006.

\bibitem{carkeetExactParametricConfidence2015}
Carkeet A.
\newblock Exact Parametric Confidence Intervals for {{Bland-Altman}} Limits of
  Agreement.
\newblock {\itshape Optometry and Vision Science: Official Publication of the
  American Academy of Optometry}. 2015;92(3):e71--80.

\bibitem{seghierImageSegmentationEvaluation2024}
Seghier ML.
\newblock Image {{Segmentation Evaluation With}} the {{Dice Index}}:
  {{Methodological Issues}}.
\newblock {\itshape International Journal of Imaging Systems and Technology}.
  2024;34(6):e23203.

\bibitem{kooGuidelineSelectingReporting2016}
Koo TK, Li MY.
\newblock A {{Guideline}} of {{Selecting}} and {{Reporting Intraclass
  Correlation Coefficients}} for {{Reliability Research}}.
\newblock {\itshape Journal of Chiropractic Medicine}. 2016;15(2):155--163.

\bibitem{monesRateProgressionGeographic2018}
Mon{\'e}s J, Biarn{\'e}s M.
\newblock The {{Rate}} of {{Progression}} of {{Geographic Atrophy Decreases
  With Increasing Baseline Lesion Size Even After}} the {{Square Root
  Transformation}}.
\newblock {\itshape Translational Vision Science \& Technology}. 2018;7(6):40.

\bibitem{yoshidaDeepLearningApproaches2025}
Yoshida K, Anegondi N, Pely A, et~al.
\newblock Deep {{Learning Approaches}} to {{Predict Geographic Atrophy
  Progression Using Three-Dimensional OCT Imaging}}.
\newblock {\itshape Translational Vision Science \& Technology}. 2025;14(2):11.

\bibitem{mishraAutomatedRetinalLayer2020}
Mishra Z, Ganegoda A, Selicha J, Wang Z, Sadda SR, Hu Z.
\newblock Automated {{Retinal Layer Segmentation Using Graph-based Algorithm
  Incorporating Deep-learning-derived Information}}.
\newblock {\itshape Scientific Reports}. 2020;10:9541.

\bibitem{ansariEvaluatingProgressionRetinal2025a}
Ansari G, Sch{\"a}rer N, Pfau K, et~al.
\newblock Evaluating the {{Progression}} of {{Retinal Sensitivity Loss}} in
  {{Geographic Atrophy Using Machine-Learning-Based Structure-Function
  Correlation}} ({{OMEGA}} 2).
\newblock {\itshape Investigative Ophthalmology \& Visual Science}.
  2025;66(11):34.

\bibitem{voglSpatiotemporalAlterationsRetinal2021a}
Vogl WD, Bogunovi{\'c} H, Waldstein SM, Riedl S, {Schmidt-Erfurth} U.
\newblock Spatio-Temporal Alterations in Retinal and Choroidal Layers in the
  Progression of Age-Related Macular Degeneration ({{AMD}}) in Optical
  Coherence Tomography.
\newblock {\itshape Scientific Reports}. 2021;11(1):5743.

\bibitem{schmidt-erfurthLongitudinalAssessmentProgressive2025}
{Schmidt-Erfurth} U, Birner K, Mai J, et~al.
\newblock Longitudinal Assessment of Progressive Functional Decline in Areas of
  Intact Retina, {{RPE}} and {{EZ}} Loss on {{OCT}} Based on Structure/Function
  Correlation in Geographic Atrophy.
\newblock {\itshape Investigative Ophthalmology \& Visual Science}.
  2025;66(8):5845--5845.

\bibitem{frank-publigQuantificationsOuterRetinal2025}
{Frank-Publig} S, Bogunovic H, Birner K, et~al.
\newblock Quantifications of {{Outer Retinal Bands}} in {{Geographic Atrophy}}
  by {{Comparing Superior Axial Resolution}} and {{Conventional OCT}}.
\newblock {\itshape Investigative Ophthalmology \& Visual Science}.
  2025;66(4):65.

\bibitem{leeChallengesAssociatedEllipsoid2021}
Lee KE, Heitkotter H, Carroll J.
\newblock Challenges {{Associated With Ellipsoid Zone Intensity Measurements
  Using Optical Coherence Tomography}}.
\newblock {\itshape Translational Vision Science \& Technology}.
  2021;10(12):27.

\end{thebibliography}

\clearpage

\section*{Abbreviations and Acronyms}
\begin{multicols}{2}
	\printacronyms[name=List of Acronyms,sort=true,heading=none]
	\onehalfspacing
\end{multicols}
\newpage

\section*{Supplemental Content: Supplemental Figures}

\captionsetup[figure]{name={Figure S},labelsep=period}

\setcounter{figure}{0}
\begin{figure}[h!]
	\centering
	\includegraphics[width=1.0\linewidth]{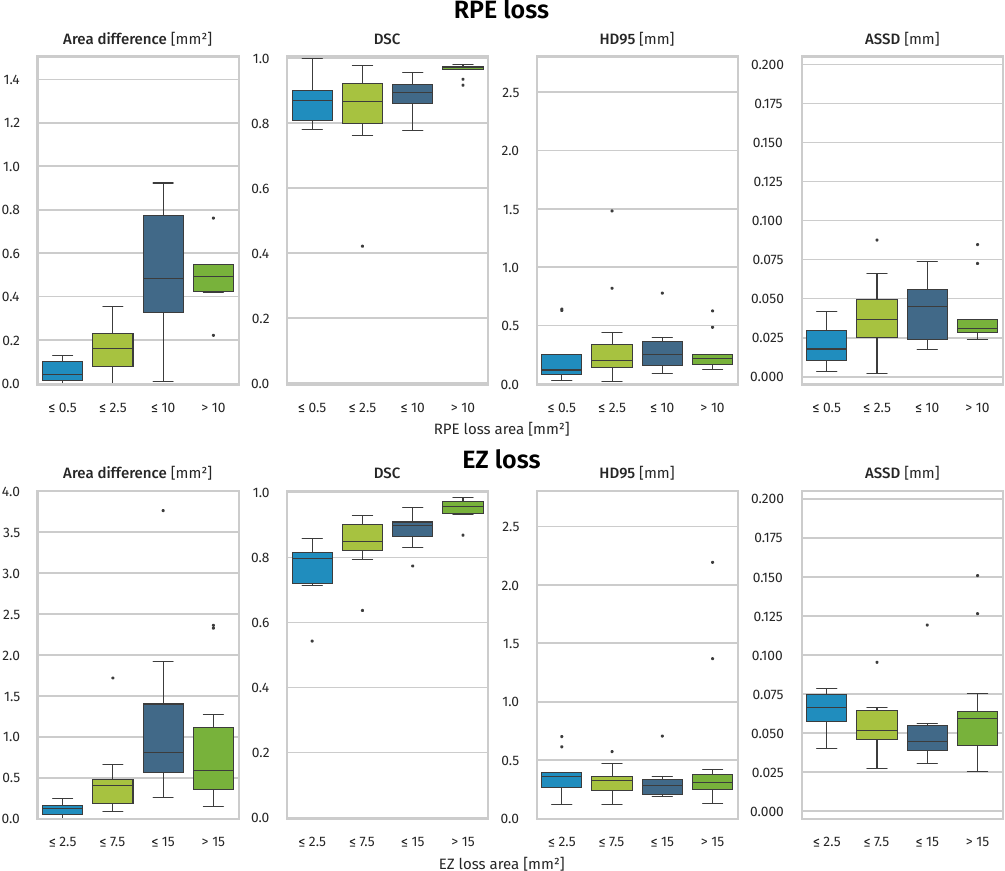}
	\caption{Segmentation performance stratified by lesion size comparing manual and automated segmentations of RPE loss (top) and EZ loss (bottom). 
    Reported metrics are the absolute difference between measured and annotated lesion size, \ac{DSC} for segmentation overlap, \acf{HD95} and \acf{ASSD} for lesion surface accuracy.}
	\label{fig:sup_stratification}
\end{figure}

\begin{figure}[h!]
	\centering
	\includegraphics[width=1.0\linewidth]{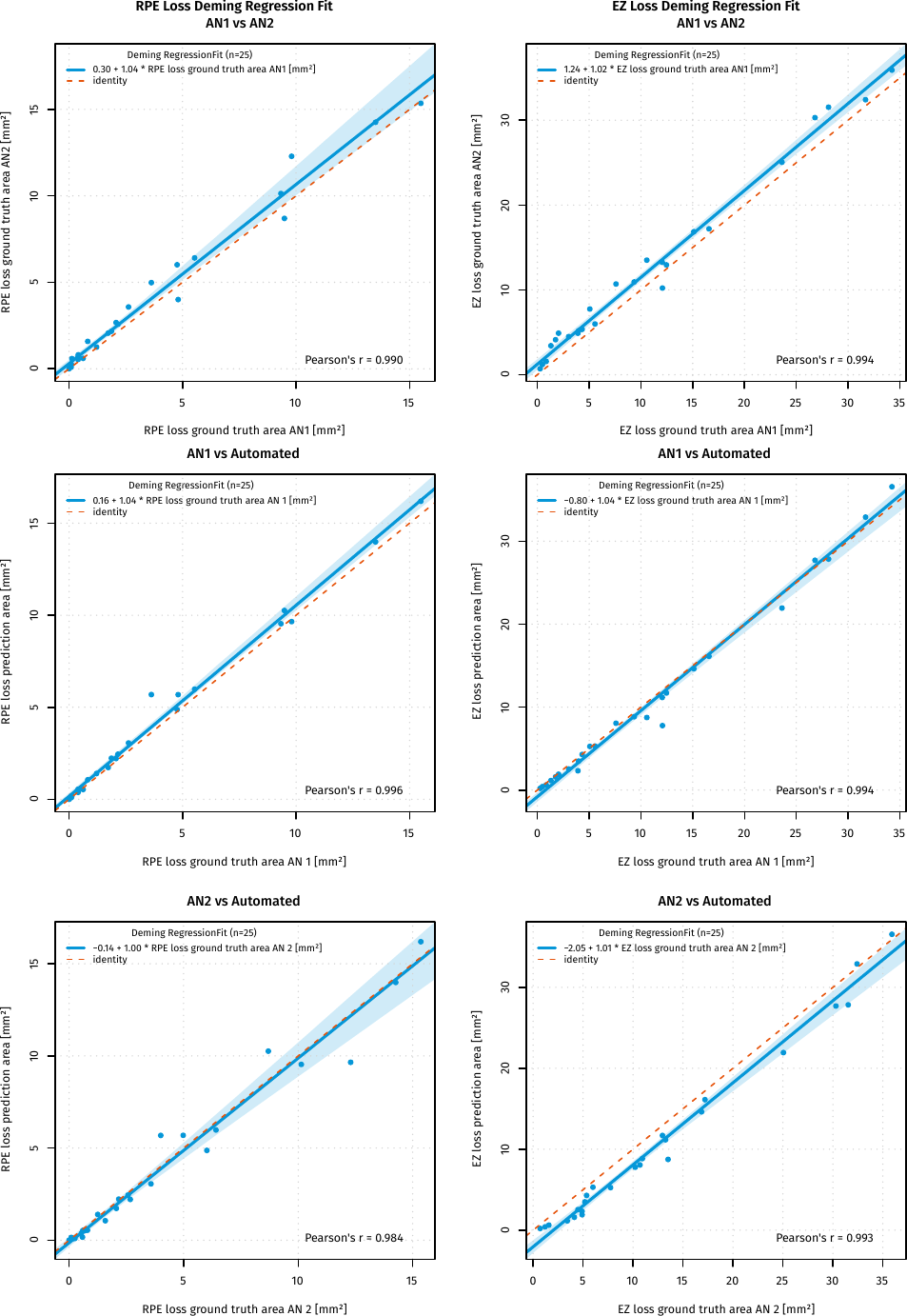}
	\caption{Deming regression fits for RPE loss area(left) and EZ loss area (right) inter-reader reliability comparing reader groups AN1 vs AN2 (top), AN1 vs automated segmentation (center) and AN2 vs automated segmentation (bottom)}
	\label{fig:sup_irr_deming}
\end{figure}

\begin{figure}[h!]
	\centering
	\includegraphics[width=1.0\linewidth]{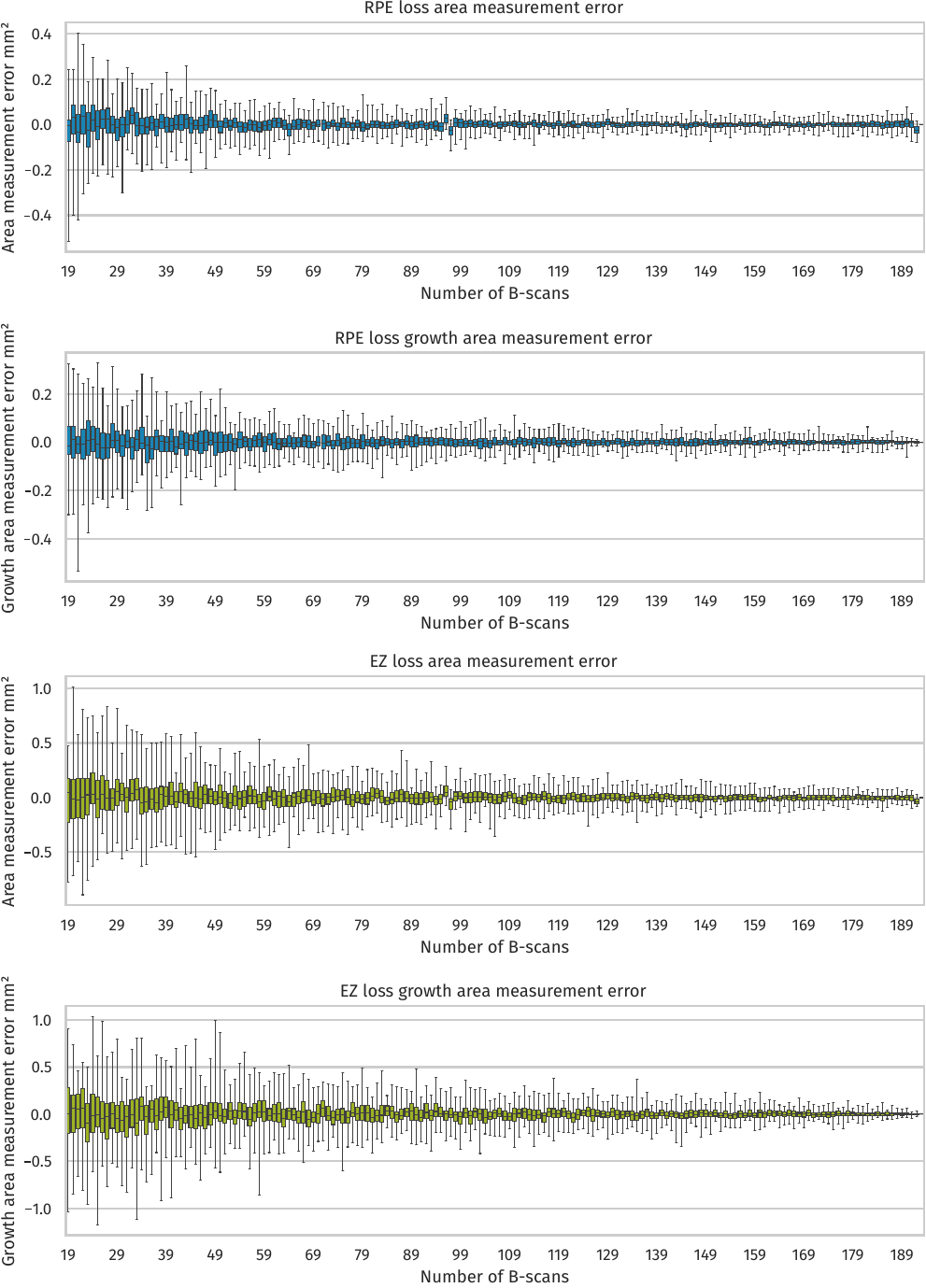}
	\caption{Effect of B-scan density on RPE loss and EZ loss measurement. Boxplot showing distribution of measurement errors of RPE loss area, RPE loss growth area growth, EZ loss areas, and EZ loss growth area (top to bottom), comparing synthetically downsampled segmentations in the range of 19 to 192 B-Scans to ground-truth segmentation with 193 B-scans. Whiskers of the boxplots indicate the full range of data points.}
	\label{fig:sup_density}
\end{figure}

\newpage
\section*{Supplemental Content: Supplemental Tables}
\captionsetup[table]{name={Table S},labelsep=period}
\setcounter{table}{0}

\begin{sidewaystable}[h!]
\caption{Comparison of the total RPE-loss (top) and EZ-loss (bottom) area measurements between manual ground truth annotation and automated segmentation; stratified by RPE loss area and EZ loss area. Owing to the small number of eyes per stratum (n = 7 to 15), confidence intervals are wide.}
\label{tab:sup_rpeloss}
\begin{footnotesize}
\begin{tabular}{@{}L{.15\linewidth}*{4}{L{.19\linewidth}}@{}}
\toprule
\textbf{Stratification RPE loss area} &
  \multicolumn{1}{c}{\textbf{$\leq 0.5$ {[}mm²{]}}} &
  \multicolumn{1}{c}{\textbf{0.5--2.5 {[}mm²{]}}} &
  \multicolumn{1}{c}{\textbf{2.5--10 {[}mm²{]}}} &
  \multicolumn{1}{c}{\textbf{$> 10$ {[}mm²{]}}} \\ \midrule
\textbf{Count}                      & \textbf{9}                    & \textbf{15}                   & \textbf{10}                  & \textbf{9}                   \\ \midrule
\textbf{Metric}                     & \multicolumn{4}{c}{\textbf{mean {[}95\% CI{]} ± std | median ± IQR}}                                                                      \\ \midrule
\textbf{AD loss area} {[}mm²{]}     & 0.06 {[}0.03 to 0.09{]} ± 0.05 |\newline 0.04 ± 0.09 & 0.17 {[}0.12 to 0.22{]} ± 0.11 |\newline 0.16 ± 0.15 & 0.61 {[}0.37 to 1.09{]} ± 0.55 |\newline 0.49 ± 0.45 & 0.60 {[}0.45 to 0.99{]} ± 0.38 |\newline 0.49 ± 0.13 \\
\textbf{PD loss area} {[}\%{]}     & 23.31 {[}16.72 to 33.01{]} ± 12.22 |\newline 23.09 ± 15.43 & 15.74 {[}10.59 to 23.80{]} ± 12.73 |\newline 16.31 ± 14.05 & 14.50 {[}8.17 to 29.85{]} ± 15.59 |\newline 9.50 ± 12.81 & 3.41 {[}2.63 to 4.39{]} ± 1.47 |\newline 3.63 ± 1.74 \\
\textbf{Area Prediction} {[}mm²{]}  & 0.29 {[}0.16 to 0.45{]} ± 0.23 |\newline 0.18 ± 0.41 & 1.31 {[}0.98 to 1.68{]} ± 0.73 |\newline 1.15 ± 1.25 & 5.54 {[}4.34 to 7.62{]} ± 2.54 |\newline 5.20 ± 2.40 & 18.11 {[}14.15 to 22.22{]} ± 6.55 |\newline 16.08 ± 11.26 \\
\textbf{Area GT} {[}mm²{]} & 0.25 {[}0.13 to 0.37{]} ± 0.19 |\newline 0.15 ± 0.32 & 1.19 {[}0.93 to 1.51{]} ± 0.62 |\newline 0.93 ± 1.11 & 4.93 {[}3.76 to 6.87{]} ± 2.47 |\newline 4.13 ± 2.10 & 17.61 {[}13.87 to 21.46{]} ± 6.22 |\newline 15.53 ± 11.07 \\
\textbf{DSC}                        & 0.87 {[}0.83 to 0.92{]} ± 0.07 |\newline 0.87 ± 0.09 & 0.84 {[}0.73 to 0.89{]} ± 0.14 |\newline 0.87 ± 0.12 & 0.88 {[}0.85 to 0.91{]} ± 0.06 |\newline 0.89 ± 0.06 & 0.96 {[}0.94 to 0.97{]} ± 0.02 |\newline 0.97 ± 0.01 \\
\textbf{HD95} {[}mm{]}                     & 0.23 {[}0.09 to 0.44{]} ± 0.25 |\newline 0.12 ± 0.17 & 0.33 {[}0.21 to 0.67{]} ± 0.37 |\newline 0.20 ± 0.19 & 0.29 {[}0.20 to 0.46{]} ± 0.20 |\newline 0.25 ± 0.21 & 0.27 {[}0.19 to 0.41{]} ± 0.17 |\newline 0.21 ± 0.08 \\
\textbf{ASSD} {[}mm{]}                      & 0.02 {[}0.01 to 0.03{]} ± 0.01 |\newline 0.02 ± 0.02 & 0.06 {[}0.03 to 0.18{]} ± 0.10 |\newline 0.04 ± 0.02 & 0.06 {[}0.04 to 0.12{]} ± 0.06 |\newline 0.05 ± 0.03 & 0.04 {[}0.03 to 0.06{]} ± 0.02 |\newline 0.03 ± 0.01 \\
\textbf{Sensitivity}                & 0.93 {[}0.90 to 0.96{]} ± 0.04 |\newline 0.93 ± 0.05 & 0.88 {[}0.72 to 0.94{]} ± 0.18 |\newline 0.96 ± 0.14 & 0.94 {[}0.90 to 0.96{]} ± 0.05 |\newline 0.96 ± 0.03 & 0.97 {[}0.94 to 0.99{]} ± 0.03 |\newline 0.99 ± 0.01 \\
\textbf{Specificity}                & 1.00 {[}1.00 to 1.00{]} ± 0.00 |\newline 1.00 ± 0.00 & 0.99 {[}0.99 to 1.00{]} ± 0.00 |\newline 0.99 ± 0.01 & 0.97 {[}0.96 to 0.98{]} ± 0.02 |\newline 0.98 ± 0.01 & 0.95 {[}0.91 to 0.97{]} ± 0.04 |\newline 0.96 ± 0.04 \\
\textbf{Precision}                  & 0.78 {[}0.73 to 0.86{]} ± 0.10 |\newline 0.77 ± 0.13 & 0.82 {[}0.77 to 0.87{]} ± 0.09 |\newline 0.81 ± 0.13 & 0.83 {[}0.77 to 0.88{]} ± 0.09 |\newline 0.84 ± 0.08 & 0.95 {[}0.94 to 0.96{]} ± 0.02 |\newline 0.96 ± 0.03 \\
\textbf{NPV}                        & 1.00 {[}1.00 to 1.00{]} ± 0.00 |\newline 1.00 ± 0.00 & 1.00 {[}0.99 to 1.00{]} ± 0.00 |\newline 1.00 ± 0.00 & 0.99 {[}0.99 to 0.99{]} ± 0.01 |\newline 0.99 ± 0.00 & 0.98 {[}0.98 to 0.99{]} ± 0.01 |\newline 0.99 ± 0.02 \\ 
\toprule
\textbf{Stratification EZ loss area} & \textbf{$\leq 2.5$ {[}mm²{]}}  & \textbf{2.5--7.5 {[}mm²{]}}  & \textbf{7.5--15 {[}mm²{]}}     & \textbf{$> 15$ {[}mm²{]}} \\ \midrule
\textbf{Count}            & \textbf{9}                 & \textbf{14}                 & \textbf{7}                  & \textbf{13}                 \\ \midrule
\textbf{Metric}           & \multicolumn{4}{c}{\textbf{mean {[}95\% CI{]} ± std | median ± IQR}}                                                               \\ \midrule
\textbf{AD loss area} {[}mm²{]}    & 0.11 {[}0.06 to 0.17{]} ± 0.09 |\newline 0.12 ± 0.12 & 0.45 {[}0.32 to 0.76{]} ± 0.41 |\newline 0.41 ± 0.28 & 1.25 {[}0.65 to 2.47{]} ± 1.23 |\newline 0.80 ± 0.84 & 0.84 {[}0.51 to 1.31{]} ± 0.74 |\newline 0.59 ± 0.76 \\
\textbf{PD loss area} {[}\%{]}     & 10.69 {[}6.30 to 15.78{]} ± 7.68 |\newline 9.60 ± 7.47 & 10.75 {[}7.09 to 18.85{]} ± 10.67 |\newline 8.58 ± 8.16 & 11.31 {[}6.26 to 21.51{]} ± 10.30 |\newline 7.15 ± 7.55 & 3.27 {[}2.14 to 5.11{]} ± 2.69 |\newline 2.24 ± 2.91 \\
\textbf{Area Prediction} {[}mm²{]} & 1.29 {[}0.85 to 1.77{]} ± 0.75 |\newline 1.16 ± 1.23 & 4.33 {[}3.60 to 5.33{]} ± 1.74 |\newline 3.95 ± 2.25 & 9.19 {[}8.15 to 10.55{]} ± 1.76 |\newline 9.05 ± 2.41 & 25.78 {[}22.41 to 28.85{]} ± 6.27 |\newline 27.42 ± 7.52 \\
\textbf{Area Groundtruth} {[}mm²{]}            & 1.40 {[}0.94 to 1.84{]} ± 0.74 |\newline 1.40 ± 1.17 & 4.67 {[}3.98 to 5.59{]} ± 1.59 |\newline 4.37 ± 1.79 & 10.44 {[}8.89 to 11.61{]} ± 1.95 |\newline 11.09 ± 2.98 & 25.83 {[}22.75 to 28.53{]} ± 5.70 |\newline 28.01 ± 5.44 \\
\textbf{DSC}              & 0.76 {[}0.69 to 0.81{]} ± 0.10 |\newline 0.80 ± 0.10 & 0.85 {[}0.80 to 0.87{]} ± 0.08 |\newline 0.85 ± 0.08 & 0.88 {[}0.83 to 0.92{]} ± 0.06 |\newline 0.90 ± 0.04 & 0.95 {[}0.93 to 0.96{]} ± 0.03 |\newline 0.96 ± 0.04 \\
\textbf{HD95} {[}mm{]}    & 0.37 {[}0.26 to 0.49{]} ± 0.19 |\newline 0.36 ± 0.13 & 0.31 {[}0.26 to 0.37{]} ± 0.12 |\newline 0.32 ± 0.11 & 0.32 {[}0.24 to 0.53{]} ± 0.18 |\newline 0.28 ± 0.13 & 0.51 {[}0.29 to 1.01{]} ± 0.59 |\newline 0.30 ± 0.13 \\
\textbf{ASSD }{[}mm{]}    & 0.06 {[}0.05 to 0.07{]} ± 0.01 |\newline 0.07 ± 0.02 & 0.05 {[}0.05 to 0.06{]} ± 0.02 |\newline 0.05 ± 0.02 & 0.05 {[}0.04 to 0.09{]} ± 0.03 |\newline 0.04 ± 0.02 & 0.06 {[}0.05 to 0.09{]} ± 0.04 |\newline 0.06 ± 0.02 \\
\textbf{Sensitivity}      & 0.72 {[}0.64 to 0.77{]} ± 0.10 |\newline 0.74 ± 0.12 & 0.81 {[}0.74 to 0.85{]} ± 0.11 |\newline 0.83 ± 0.12 & 0.84 {[}0.75 to 0.88{]} ± 0.10 |\newline 0.88 ± 0.07 & 0.95 {[}0.93 to 0.96{]} ± 0.03 |\newline 0.97 ± 0.05 \\
\textbf{Specificity}      & 0.99 {[}0.99 to 1.00{]} ± 0.01 |\newline 0.99 ± 0.01 & 0.98 {[}0.97 to 0.99{]} ± 0.01 |\newline 0.99 ± 0.01 & 0.98 {[}0.97 to 0.99{]} ± 0.01 |\newline 0.98 ± 0.01 & 0.89 {[}0.83 to 0.93{]} ± 0.09 |\newline 0.93 ± 0.11 \\
\textbf{Precision}        & 0.81 {[}0.74 to 0.86{]} ± 0.10 |\newline 0.81 ± 0.13 & 0.89 {[}0.86 to 0.91{]} ± 0.05 |\newline 0.90 ± 0.06 & 0.94 {[}0.93 to 0.96{]} ± 0.02 |\newline 0.94 ± 0.03 & 0.95 {[}0.92 to 0.97{]} ± 0.04 |\newline 0.97 ± 0.04 \\
\textbf{NPV}              & 0.99 {[}0.99 to 0.99{]} ± 0.01 |\newline 0.99 ± 0.01 & 0.97 {[}0.96 to 0.98{]} ± 0.01 |\newline 0.97 ± 0.02 & 0.93 {[}0.89 to 0.96{]} ± 0.04 |\newline 0.96 ± 0.05 & 0.89 {[}0.84 to 0.92{]} ± 0.07 |\newline 0.89 ± 0.06 \\ \bottomrule
\end{tabular}
\end{footnotesize}

\scriptsize{\Acf{AD}, \acf{PD}, \acf{DR}, \acf{GT}, \acf{AD}, \acf{PD}, \acf{DSC}, \acf{NPV}, \acf{HD95}, \acf{ASSD}}

\end{sidewaystable}

\begin{table}[h!]
\caption{EZ loss / RPE loss area ratio split into 4 quartiles. Areas are based on segmentation measurements of all Spectralis scans of the OAKS \& DERBY study eyes. 7 scans were removed for analysis due to image quality issues. Q1 is associated with slowest progression rate and Q4 with fastest progression rate on average. }
\label{tab:ez_rpe_ratio}
\begin{tabular}{@{}lll@{}}
\toprule
\textbf{Total number of scans} &
  \textbf{Scans used for analysis} &
  \textbf{EZ/RPE loss ratio quartiles} \\ \midrule
904 &
  897 &
  \begin{tabular}[c]{@{}l@{}}
  Q1 $< 1.37$; \\ Q2 $\geq 1.37$ to $< 1.67$; \\ Q3 $\geq 1.67$ to $< 2.17$; \\ Q4  $\geq 2.17$\end{tabular} \\ \bottomrule
\end{tabular}
\end{table}

\end{document}